\documentclass[journal]{IEEEtran}
\usepackage{cite}

\ifCLASSINFOpdf
\else
\fi
\usepackage{amsmath}
\usepackage{enumerate}

\usepackage{url}

\usepackage{booktabs}
\usepackage{graphicx}
\usepackage[caption=false]{subfig}
\usepackage{ulem}
\usepackage{tikz}
\usetikzlibrary{shapes,arrows,positioning}
\usepackage{xcolor}
 
\newcommand{\lastrev}[1]{{\color{black}{{#1}}}}

\begin{document}
%
\title{3D Reconstruction in Canonical Co-ordinate Space from Arbitrarily Oriented 2D Images 
	}
%
%
%

\author{
    Benjamin Hou, Bishesh Khanal, Amir Alansary, Steven McDonagh, Alice Davidson, Mary Rutherford, \\ Jo V. Hajnal, Daniel Rueckert, Ben Glocker and Bernhard Kainz
}

\markboth{SUBMITTED TO IEEE TRANSACTIONS ON MEDICAL IMAGING}%
{Shell \MakeLowercase{\textit{et al.}}: Bare Demo of IEEEtran.cls for IEEE Journals}
%




\maketitle

\begin{abstract}
Limited capture range, and the requirement to provide high quality initialization for optimization-based 2D/3D image registration methods, can significantly degrade the performance of 3D image reconstruction and motion compensation pipelines. Challenging clinical imaging scenarios, which contain significant subject motion such as fetal in-utero imaging, complicate the 3D image and volume reconstruction process.

In this paper we present a learning based image registration method capable of predicting 3D rigid transformations of arbitrarily oriented 2D image slices, with respect to a learned canonical atlas co-ordinate system. Only image slice intensity information is used to perform registration and canonical alignment, no spatial transform initialization is required. To find image transformations we utilize a Convolutional Neural Network (CNN) architecture to learn the regression function capable of mapping 2D image slices to a 3D canonical atlas space.

We extensively evaluate the effectiveness of our approach quantitatively on simulated Magnetic Resonance Imaging (MRI), fetal brain imagery with synthetic motion and further demonstrate qualitative results on real fetal MRI data where our method is integrated into a full reconstruction and motion compensation pipeline. Our learning based registration achieves an average spatial prediction error of 7 mm on simulated data and produces qualitatively improved reconstructions for heavily moving fetuses with gestational ages of approximately 20 weeks. Our model provides a general and computationally efficient solution to the 2D/3D registration initialization problem and is suitable for real-time scenarios.
\end{abstract}


%
\IEEEpeerreviewmaketitle

\section{Introduction}
%
%
%
%
\IEEEPARstart{R}{econstructing} a 3D volume from misaligned and motion corrupted 2D images is a challenging task. The process involves labor intensive pre-processing steps, such as manual landmark matching, exhibiting both inter and intra-observer variance. Pre-processing is a necessary step to achieve acceptable input for  intensity-based pose optimization for a volume reconstruction process. Optimization facilitates alignment and combination of intensity data from multiple image sources into a common co-ordinate system. 


Image registration is also required for applications such as atlas-based segmentation~\cite{Aljabar2009726}, tracking~\cite{Miao2016}, image fusion from multiple modalities~\cite{987693} and clinical analysis of images visualized in an anatomical standard co-ordinate system~\cite{brainatlas}. All of these applications suffer from poor initialization of automatic registration methods, which must be alleviated by manual pre-processing.

For the 2D/3D case, two distinct registration strategies can be categorized as volume-to-slice and slice-to-volume techniques. The former is concerned with aligning a volume to a given image, \emph{e.g.}, aligning an intra-operative C-Arm X-Ray image to a pre-operative volumetric scan. In contrast, the latter is concerned with aligning multiple misaligned 2D slices into a unique co-ordinate system of a reference volume. A recent review of slice-to-volume registration techniques is given in~\cite{Ferrante2017}.

Arbitrary subject motion can invalidate slice alignment assumptions that are based on the scanner co-ordinate system, and manual intervention may be necessary. Manual correction of slice-to-volume registration often becomes unfeasible in practice due to the magnitude of image data involved. 
Manual volume-to-slice registration is often easier to achieve, since manual alignment of one or a few 3D volumes to a single 2D slice or projection is less time consuming than manual alignment of hundreds of individual slices into a common co-ordinate system. Landmark-based techniques can help to automate this process, but this approach is heavily dependent on detection accuracy and robustness of the calculated homography between locations and the descriptive power of the used landmark encoding. 2D slices also do not provide the required 3D information to establish robust landmark matching, therefore this technique cannot be used on applications such as motion compensation in fetal imaging. 
For slice-to-volume registration, methods such as~\cite{rousseau2006registration,jiang2007mri,gholipour2010robust,kim2010intersection,murgasova2012reconstruction,kainz2015fast,Alansary2016} are effective in cases where a coarsely aligned initialization of the 3D volume is provided to initialize the reconstruction process. This initial 3D reference volume is used as a 2D/3D registration target to seed the iterative estimation of the slice orientation and intensity data combination. Reasonably good initial coarse alignment of 2D image slices is critical to form the seed reference volume. 

Traditional intensity-based slice-to-volume reconstruction methods~\cite{gholipour2010robust,murgasova2012reconstruction} involve solving the inverse problem of  super-resolution from slice acquisitions~\cite{park2003super}, shown in Eq.~\ref{eqn:slice-acquisition-model}:

\begin{equation}
    y_i = D_i B_i S_i M_i x + n_i; \quad i = 1, 2, \dots , N
\label{eqn:slice-acquisition-model}
\end{equation} 

$y_i$ denotes the $i$th low resolution image obtained during scan time, $D_i$ is the down-sampling matrix, $B_i$ is the blurring matrix, $S_i$ is the slice selection matrix, $M_i$ is a matrix of motion parameters, and $x$ is the high resolution 3D volume with $n_i$ as a noise vector. More commonly, $D_i$, $B_i$ and $S_i$ are grouped together in a single matrix $W_i$. Obtaining the true high-resolution volume $x$ is ill-posed, as it requires inversion of the large, ill-defined matrix $W_i$.


Alternatively, various optimization methods~\cite{gholipour2010robust,murgasova2012reconstruction,kainz2015fast} have been applied to obtain a good approximation of the true volume $x$. These are typically two step iterative methods, consisting of Slice-to-Volume Registration (SVR) and Super Resolution (SR). In each iteration, each slice is registered to a target volume, followed by super resolution reconstruction. The output volume is therefore used as a registration target for the next iteration. Hence, a good initial alignment is crucial. If any slice cannot be registered, it is discarded and not further utilized for reconstruction. 

The optimization methods employed in this domain typically do not guarantee a globally optimal registration solution from arbitrarily seeded slice alignment. The function that maps each 2D slice to its correct anatomical position in 3D space may be subject to local minima and the requirement for small initial misalignment typically improves result quality. Previous work have attempted to make this optimization robust by introducing appropriate objective functions and outlier rejection strategies based on robust statistics~\cite{gholipour2010robust,murgasova2012reconstruction}. Despite these efforts, good reconstruction quality still depends on having good initial alignment of adjacent and intersecting slices.

The robustness of (semi-)automatic 2D/3D registration methods is characterized by their \emph{capture range}, which is the maximum transformation misalignment from which a method can recover good spatial alignment. When the data available is limited, as common for the 2D/3D case, this task becomes very challenging.



We provide a solution for the 2D/3D capture range problem, which is applicable to many medical imaging scenarios and can be used with any registration method. We demonstrate the capabilities of the method in the current work using in-utero fetal MRI data as a working example. During gestation, high-quality and high-resolution stacks of slices can be acquired through fast scanning techniques such as Single Shot Fast Spin Echo (ssFSE)~\cite{jiang2007mri}. Slices can be obtained in a fraction of a second, thus freezing  in-plane motion. Random motion (\emph{e.g.}, a fetus that is awake and active during a scan) or oscillatory motion (\emph{e.g.}, maternal breathing), is likely to cause adjacent slices to become incoherent and corrupt a 3D scan volume. State-of-the-art slice-to-volume registration methods~\cite{murgasova2012reconstruction,kainz2015fast,Alansary2016} are able to compensate for this motion and to reconstruct a consistent volume from overlapping motion-corrupted orthogonal stacks of 2D slices. These methods tend to fail for volumes with large initial misalignment between the 2D input slices. Thus, SVR works best for neonates and older fetuses who have less space to move. 
However, for early diagnostics, detailed anatomical reconstructions are required from an early age (younger than 25 weeks). 


\subsection{Related Work}\label{sec:relatedworks}
\noindent \textbf{2D/3D registration:} Image registration methods that can compensate for large initial alignment offsets usually require robust automatic or manual anatomical landmark detection~\cite{Johnson2002,Markelj2012,Oliveira2014} with subsequent 3D location matching. This often relies on the use of fiducial markers~\cite{Penney1998,Susil2006,Kainz2008,Oliveira2014} involving special equipment and/or invasive procedures. 
Manual annotation of landmarks by a domain expert is the current clinical practice to initialize automatic image registration~\cite{Oliveira2014}. Fully automatic methods are difficult to integrate into the clinical workflow because of their limited reliability, complex nature, long computation times, susceptibility to error and lack of generalization. 

Miao et al.~\cite{Miao2016} use Convolutional Neural Networks (CNNs) to automatically estimate the spatial arrangement of landmarks in projection images. Their method utilizes a CNN to regress transformation residuals, $\delta t$, which refines the required transformation to register a source volume to a target X-ray image from an initially assumed position $t_n = t_{ini}$. Registration is then performed iteratively using synthetic Digitally Reconstructed Radiography (DRR) images generated from the source volume using $t_{n+1} = t_n+\delta t$.
To address inaccurate transformation mappings caused by direct regression of transformation parameters, Miao et al. train their CNN using Pose-index features (landmarks) extracted from source and target image pairs and learn their $\delta t$. Pose-index features are insensitive to transform parameters $t$ yet sensitive to change in $\delta t$. This insensitivity to $t$ can be expressed as $X(t_1,I_{t1+\delta t}) \approx X(t_2,I_{t2+\delta t}) \enskip \forall (t_1,t_2)$. The method requires a robust landmark detection algorithm, which is domain and scanner specific and the detection quality degrades for motion-corrupted data.

Similarly, Simonovsky et al.~\cite{Simonovsky2016} use CNNs to perform 2D to 3D registration between a target 2D slice and a source 3D reference volume. Instead of regressing on transformation residuals, the authors regress to a scalar that estimates the dissimilarity between two image patches, and leverage the error from back-propagation to update the transformation parameters. To this end, all operations can be efficiently computed on a GPU for high throughput.

Pei et al. \cite{Pei2017}  trained CNNs to regress the transformation parameters for 2D to 3D registration. Their regression is performed directly on image slices without feature extraction. The input to the network are pairs of target 2D X-ray images with synthetic DRR source images that are generated from a Cone-Beam Computed Tomography (CBCT) volume. Each image pair is augmented with varying levels of anisotropic diffusion. Iterative updates of the CNN yield new transformation parameters to generate new DRR images until similarly to the target X-Ray converges.


\noindent \textbf{Motion compensation:} Fetal brain MRI is a field requiring motion compensation to gain diagnostically useful 3D volumes.
Most algorithms use general 2D to 3D registration methods via gradient descent optimization over an intensity based cost function, in conjunction with a Super Resolution step to recreate the output volume. 

The earliest framework for fetal MRI reconstruction was developed by Rousseau et al.~\cite{rousseau2006registration}. It introduced steps for motion correction and 3D volume reconstruction via Scattered Data Interpolation (SDI). For slice registration, Rousseau used a gradient ascent method to maximize the Normalized Mutual Information (NMI). This was improved by Jiang et al.~\cite{jiang2007mri} who proposed to use Cross Correlation (CC) as the cost function to optimize. Jiang also proposed to over-sample the Region of Interest (RoI) with thin slices to ensure sufficient sampling. 
Gholipour et al. \cite{gholipour2010robust} integrated the mathematical model of Super Resolution in Eq.~\ref{eqn:slice-acquisition-model} into the slice-to-volume reconstruction pipeline and introduced outlier rejection. This was further improved by Murgasova et al. \cite{murgasova2012reconstruction} adding intensity matching and bias correction, as well as an EM-based model for outlier detection. 


The 2D/3D registration methods, by \cite{Miao2016}, \cite{Simonovsky2016} and \cite{Pei2017}, use CNNs to compute the unknown transformation of a given 2D slice with respect to a reference 3D volume while
\cite{rousseau2006registration,jiang2007mri,gholipour2010robust,murgasova2012reconstruction,fogtmann2012unified,TOURBIER2015584,kainz2015fast} 
use general registration algorithms to register slices to an initial 3D target. 
In both cases, a motion free reference/initial volume is required for successful 2D/3D registration, which is not guaranteed to be obtainable from a clinical scan  due to unpredictable subject motion.

Kim et al.~\cite{kim2010intersection} proposed 
to perform slice-to-volume registration by minimizing the energy of weighted Mean Square Difference (wMSD) of all slice intersection intensity profiles. This method does not require an initial target registration volume nor intermediate 3D reconstructions.
The authors were able to ``recover motion up to 15 mm of translation and 30$^\circ$ of rotation for individual slices". 

Our method also estimates slice motion without the need for volume reconstruction. However, we focus on tackling the problem of a reasonable initial slice alignment in 3D canonical space, which is not guaranteed in real scan scenarios. 
This goal is related to the natural image processing work of Kendall et al.~\cite{Kendall_2015_ICCV} who proposed PoseNet, which regresses 6-DoF camera poses from single RGB images. PoseNet is trained from RGB images taken from a particular scene, \emph{e.g.}, on an outdoor street or in a corridor. The CNN is then used to infer localization within the learned 3D space. Expanding on this idea, for a given 2D image slice, we would like to infer pose, relative to a 3D atlas space, without knowing any initial information besides the image intensities. 


Hou et al.~\cite{Hou2017} demonstrated the potential of CNNs for tackling the volume initialization problem for slice-to-volume 3D image reconstruction. The network architecture in~\cite{Hou2017} showed promising results, initializing scan slices for fetal brain in-utero volume reconstruction and for pose estimation of DRR scan images. However, it does not provide means for estimating incorrect predictions and outlier rejection. Failing to account for grossly misaligned slices, that constitute outlying samples, hinders reconstruction performance and may result in volume reconstruction failure. 
We extend~\cite{Hou2017} by rigorous evaluation of several network architectures and introduce Monte Carlo Dropout~\cite{pmlr-v48-gal16} for the purpose of establishing a prediction confidence metric.





\subsection{Contributions}
In this paper, we introduce a learning based approach that automatically learns the slice transform model of arbitrarily sampled slices relative to a canonical co-ordinate system (\emph{i.e.}, our approach learns the mapping function that maps each slice to a volumetric atlas). This is achieved using only the intensity information encoded in the slice without relying on image transformations in scanner co-ordinates. Our CNN predicts 3D rigid transformations, which are elements of a Special Euclidean group SE(3). 
Predicting canonical orientations for each slice in a collection of 3D stacks covering a RoI provides an accurate initialization for subsequent automatic 3D reconstruction and registration refinement using intensity-based optimization methods. 
Recent statistical analysis and metrics~\cite{Miolane2013}, specific to Lie groups, are incorporated to give a more accurate measure of the misalignment between a predicted slice and the corresponding ground truth. This is combined with traditional image similarity metrics such as Cross Correlation and Structural Similarity.

We report quantitative comparisons, evaluating the predictive performance of several CNN architectures with real and synthetic 2D slices that are corrupted with extreme motion. Synthetic slices, with known ground truth locations, are extracted from 3D MRI fetal brain volumes of approximately 20 weeks Gestational Age (GA). We additionally evaluate the approach qualitatively on heavily motion corrupted fetal MRI where ground truth slice transformations and/or 3D volumes are not available. By providing 2D slices that are canonically aligned to initialize a subsequent reconstruction step, we can qualitatively assess the improvement our method provides for the volume reconstruction task. 

We implement Monte Carlo dropout sampling during inference to consider the model's epistemic uncertainty, and provide a prediction confidence for each slice. This is used as a metric for outlier rejection (\emph{i.e.}, if the model is not confident about a prediction, then it can be discarded before subsequent use during 3D reconstruction).

Our approach can also be generalized to 3D-3D volumetric registration by predicting the transformation parameters of a few selected slices to be used as landmark markers. This is demonstrated in~\cite{Hou2017} by predicting thorax phantom slices where no organ specific segmentation is performed. It is also applicable to projective 2D images, which is highly valuable for X-Ray/CT registration.

\section{Method}

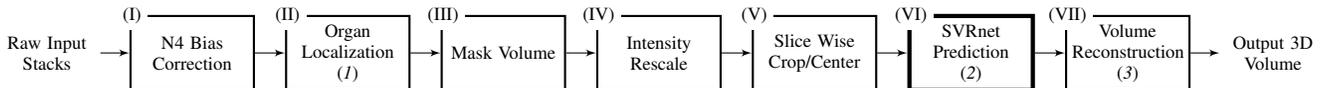
\begin {figure*}[!hbtp]
\centering
\begin{tikzpicture}[auto,
    block/.style ={rectangle, draw=black, thick, fill=white,
      text width=6em, text centered,
      minimum height=4em, node font=\scriptsize},
    block_num/.style ={fill=white,above,
      xshift=-0.8cm,yshift=0.3cm, 
      node font=\scriptsize},
    block_bold/.style ={rectangle, draw=black, ultra thick, fill=white,
      text width=6em, text centered,
      minimum height=4em, node font=\scriptsize},
    block_dash/.style ={rectangle, draw=gray, dashed, fill=white,
      text width=6em, text centered,
      minimum height=4em, node font=\scriptsize},
    block_noborder/.style ={rectangle, draw=none, thick, fill=none,
      text width=5em, text centered, minimum height=1em, node font=\scriptsize},
    line/.style ={draw, solid, -latex', shorten >=0pt},
    line_dashed/.style ={draw, dashed, -latex', shorten >=0pt}]
    \matrix [column sep=4mm,row sep=3mm] {
      \node [block_noborder] (a) {Raw Input Stacks}; 
      & \node [block] (b) {N4 Bias Correction}; 
      & \node [block] (c) {Organ Localization (\textit{1})}; 
      & \node [block] (d) {Mask Volume}; 
      & \node [block] (e) {Intensity Rescale }; 
      & \node [block] (f) {Slice Wise Crop/Center}; 
      & \node [block_bold] (g) {SVRnet Prediction \\(\textit{2})}; 
      & \node [block] (h) {Volume\\ Reconstruction (\textit{3})}; 
      & \node [block_noborder] (i) {Output 3D Volume}; \\
    };

    \draw[] (b) node[block_num] {(I)};
    \draw[] (c) node[block_num] {(II)};
    \draw[] (d) node[block_num] {(III)};
    \draw[] (e) node[block_num] {(IV)};
    \draw[] (f) node[block_num] {(V)};
    \draw[] (g) node[block_num] {(VI)};
    \draw[] (h) node[block_num] {(VII)};

    \begin{scope}[every path/.style=line]
      \path (a)   -- (b);
      \path (b)   -- (c);
      \path [line] (c) -- (d);
      \path (d)   -- (e);
      \path (e)   -- (f);
      \path (f)   -- (g);
      \path (g)   -- (h);
      \path (h)   -- (i);
      
    \end{scope}
\end{tikzpicture}
\caption{Pipeline to reconstruct volume from stacks of heavily motion corrupted scans. N.B. this is only used for test cases. (I) N4 Bias Correction is performed directly on the raw scan volume (II) \cite{rajchl2016learning,rajchl2016learning2} is used for Approximate Organ Localization to create segmentation masks for desired RoI. (III) The bias corrected volume is masked. (IV) The volume is intensity rescaled to 0~255 to match SVRnet training parameters. (V) Each slice within the volume is cropped and centered positioned on a slice for SVRnet inference. (VI) Each slice is inferred through SVRnet to obtain prediction of atlas space location. (VII) OPTIONAL: Use predicted transformations, with original or inference slice, to initialize traditional SVR algorithm(s) for further registration refinement.}
\label{fig:full-pipeline}
\end {figure*}
To fully evaluate and assess the performance of 2D/3D registration via a learning based approach, we incorporated it into a full 3D reconstruction pipeline as shown in Fig.~\ref{fig:full-pipeline}. This features three modular components: (\textit{1}) approximate organ localization, (\textit{2}) canonical slice orientation estimation, and (\textit{3}) intensity-based 3D volume reconstruction.
Organ localization (\textit{1}) is concerned with localization of a learned RoI. This can be achieved through rough manual segmentation, organ focused scan sequences or automatic methods~\cite{Konukoglu2013,keraudren2014automated,Keraudren2015}.

For 3D volume reconstruction (\textit{3}) we use a modified iterative SVR method~\cite{kainz2015fast}, incorporating super-resolution image reconstruction techniques~\cite{park2003super,kainz2015fast,Alansary2016}. This additionally allows for compensation of any remaining small misalignments between single slices, caused by prediction inaccuracies.
To provide a sufficiently high number of samples for SVR, multiple stacks of 2D-slices are acquired, ideally in orthogonal orientations~\cite{gholipour2010maximum}. We modified~\cite{kainz2015fast}; such that instead of generating an initial reference volume from all slices oriented in scanner co-ordinates, the acquired slice orientations are replaced with predicted canonical atlas co-ordinates and the iterative intensity-based reconstruction process continues from that point.
Since (\textit{1}) and (\textit{3}) can be achieved using state-of-the-art techniques, we focus in the remainder of this section on the canonical slice orientation estimation (\textit{2}), specifically the learning and prediction of 3D rigid transformations using a variety of network architectures.


At its core, our method uses a CNN, called \emph{SVRnet}, to regress transformation parameters $\hat{T_i}$, such that $\hat{T_i} = \psi(\omega_i,\Theta)$. $\Theta$ are the learned network parameters and $\omega_i$ is a 2D image slice of size $L \times L$, extracted from a series of slices acquired from a 3D volume $\Omega$. Each $\Omega$ encloses a desired RoI, organ or particular use case scenario, such that $\omega_i \in \Omega$.

To train SVRnet, slices of varying orientations and their corresponding ground truth locations are obtained from existing organ atlases or from collections of motion-free 3D volumes, \emph{e.g.}, pre-interventional scans or successfully (partially manually) motion compensated subjects. 

\subsection{Rigid Body Transformation Parameterization}\label{sec:RigidBodyTransParam}
The motion of a rigid body in 3D has six Degrees of Freedom (6 DoF). One common parameterization for this motion defines three parameters for translation ($T_x$, $T_y$, $T_z$) and three for rotation ($R_x$, $R_y$, $R_z$). To model the movement of each slice in 3D space, we divide the parameters into two categories; in-plane transformation $T_x$, $T_y$ and $R_z$ and out-of-plane transformation $T_z$, $R_x$ and $R_y$ (see Fig.~\ref{fig:samplingschema}(d-j)). If each DoF is allowed ten interval delineation, this would result in $10^{6}$ slices per organ volume. 

Automatic segmentation methods, such as \cite{rajchl2016learning,rajchl2016learning2,2017arXiv171009338S}, 
define the RoI on a slice by slice basis throughout the 3D volume. The desired RoI (\emph{e.g.}, segmented brain) is masked and center aligned within $\omega_i$, which vastly decreases the valid range for in-plane motion parameters $T_x$ and $T_y$. Similar to~\cite{Miao2016}, we additionally reduce the number of slices required to create training and validation data sets by simplifying the sample space such that it is constrained by the parameters: $T_z$, $R_x$, $R_y$ and $R_z$. We can further discount a portion of slices that yield little or no content at the extremities of the $T_z$ range, in the considered volume. 

\subsection{Data Pre-processing}
During scan-time, image intensity ranges are influenced by RoI structure and/or set by the radiologists based on visual appeal for diagnostic purposes. This causes each scanned volume to be biased differently with an intensity range that varies from scan to scan. 

Pre-processing image intensities, via Min-Max normalization or Z-score normalization, is typically a necessary step when training CNNs. The process helps to keep image features on a common scale and keep similar features across different images consistent. Z-score normalization scales all volumes to zero mean with a standard deviation of one, and is a common pre-processing step for K-Nearest Neighbor based techniques and clustering algorithms. Alternatively, a quicker approximation can be made by performing Min-Max normalization. Intensity normalization of the pixels, as shown in Fig.~\ref{fig:full-pipeline}, is performed after the RoI is masked.

For training and validation data sets, we extract $\omega_i$ from 3D motion corrected and segmented fetal brain volumes that are registered to a canonical atlas space. Fig.~\ref{fig:samplingschema}a and \ref{fig:samplingschema}b shows an example of slices, $\omega_i$, being extracted from a brain volume $\Omega$. SVR~\cite{kainz2015fast} is performed on raw scan volumes with a mask applied on fetal brain as the desired RoI. These volumes featured little fetal and maternal motion, and hence, reconstruction were successful. 3D reconstructions are intensity rescaled to 0-255 with an isotropic spacing of 0.75 $\times$ 0.75 $\times$ 0.75 mm, and further manually aligned to canonical atlas co-ordinates~\cite{brainatlas}. The resulting volume of size $L \times L \times L$ encloses a brain atlas that is center-aligned. 

For inference on raw scan data, raw volumes are N4 bias corrected first (Fig.~\ref{fig:full-pipeline}(I)). This ensures that intensities in regions affected by small  magnetic field inhomogeneities are corrected.
After RoI localization (Fig.~\ref{fig:full-pipeline}(II)) the volume is masked (Fig.~\ref{fig:full-pipeline}(III)) and intensity normalized to 0-255 (Fig.~\ref{fig:full-pipeline}(IV)). To predict the transformation parameters with SVRnet, each slice in the 3D volume is individually scaled back to isotropic spacing with the masked RoI centerd within $\omega_i$ (Fig.~\ref{fig:full-pipeline}(V)).

\subsection{Generating Training and Validation Data}\label{sec:gentraindata}

\begin{figure}[htb]
  \centering
  \subfloat[Front View]
  {\includegraphics[trim={4.5cm 2.5cm 6cm 3.5cm},clip,width=2.8cm]{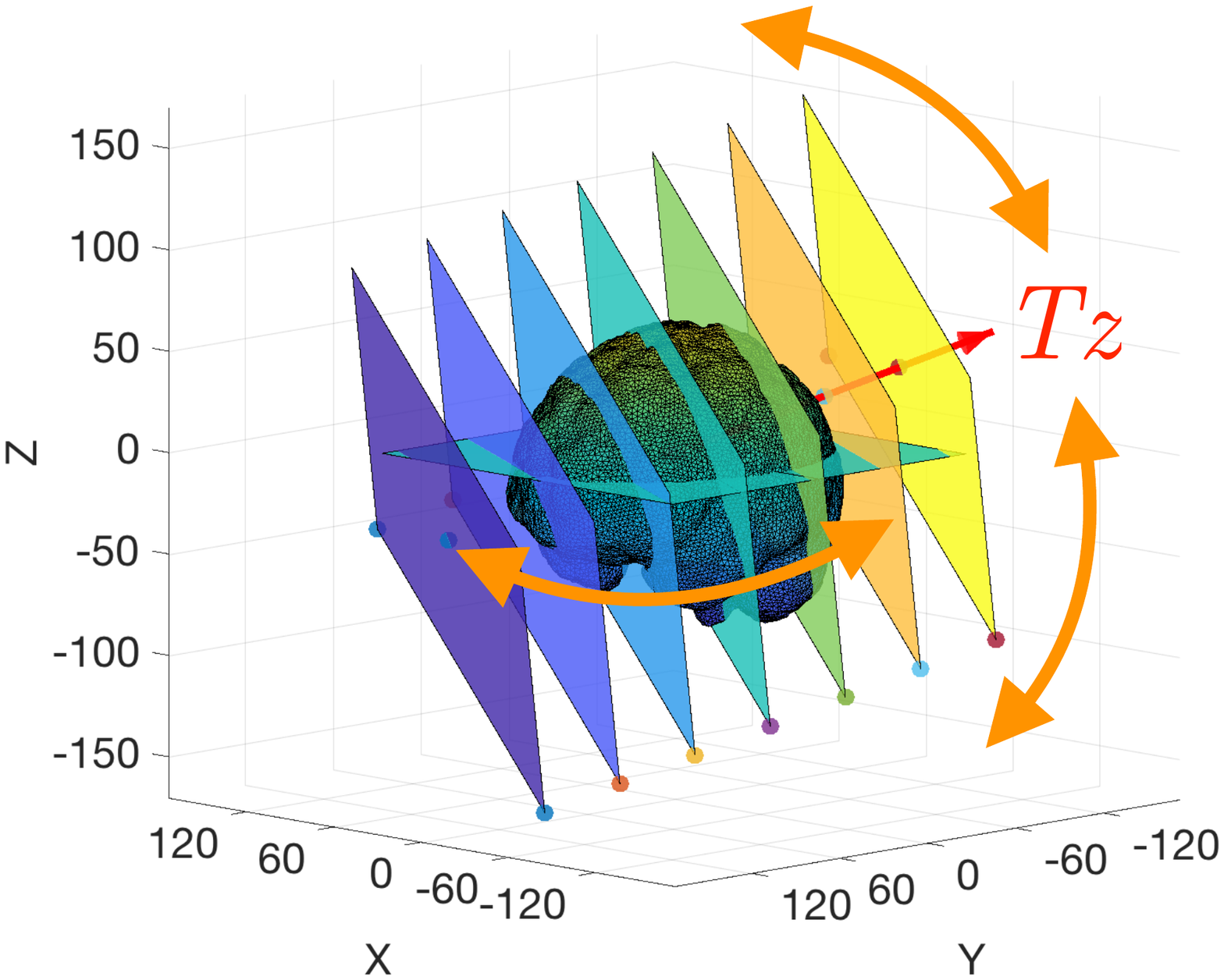}\label{fig:samplingschema_f}} \hfill 
  \subfloat[Rear View]
  {\includegraphics[trim={4.5cm 2.5cm 6cm 3.5cm},clip,width=2.8cm]{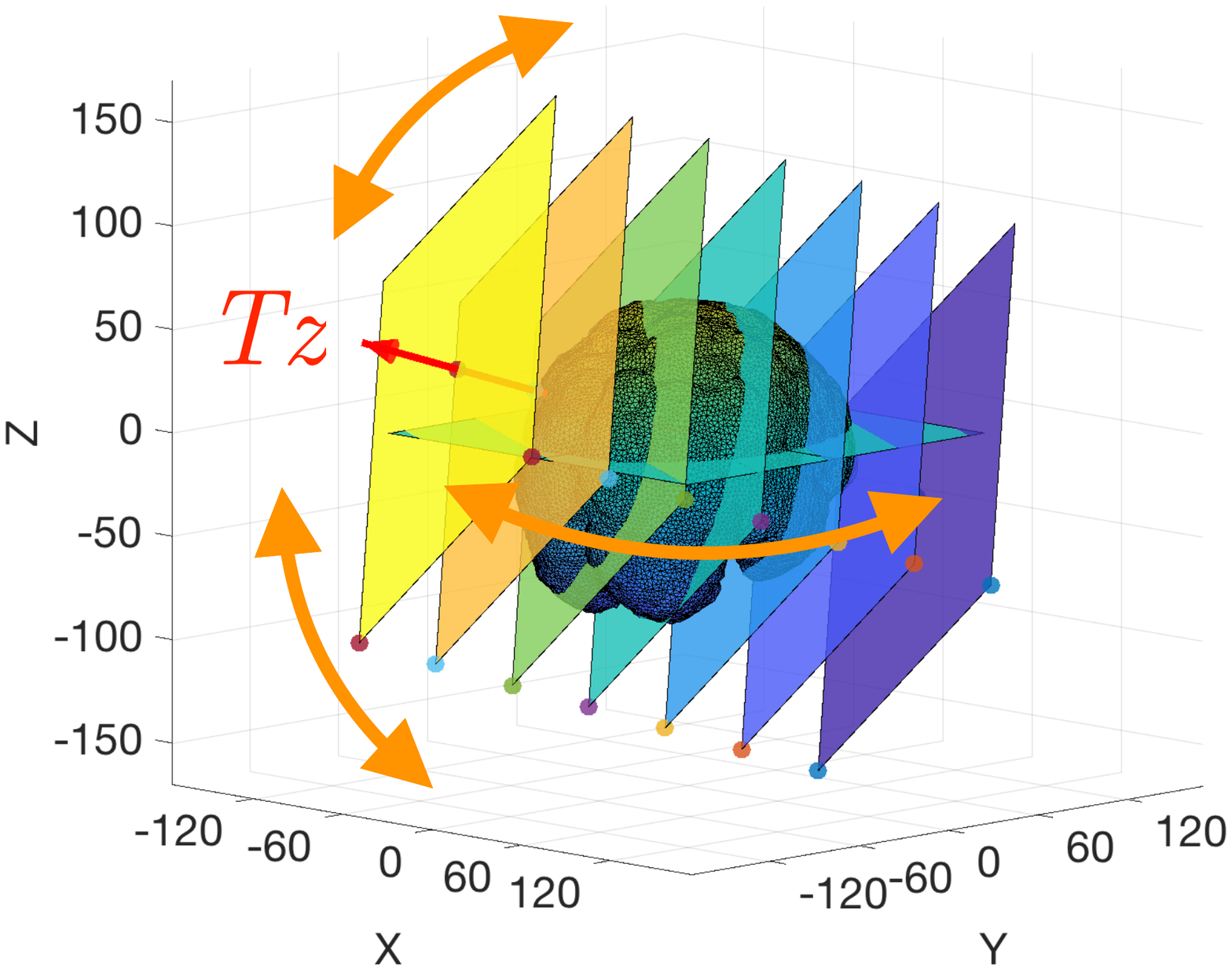}\label{fig:samplingschema_r}}  \hfill 
  \subfloat[Anchor Points]
  {\includegraphics[trim={0 -1cm 0 0},clip,width=3cm]{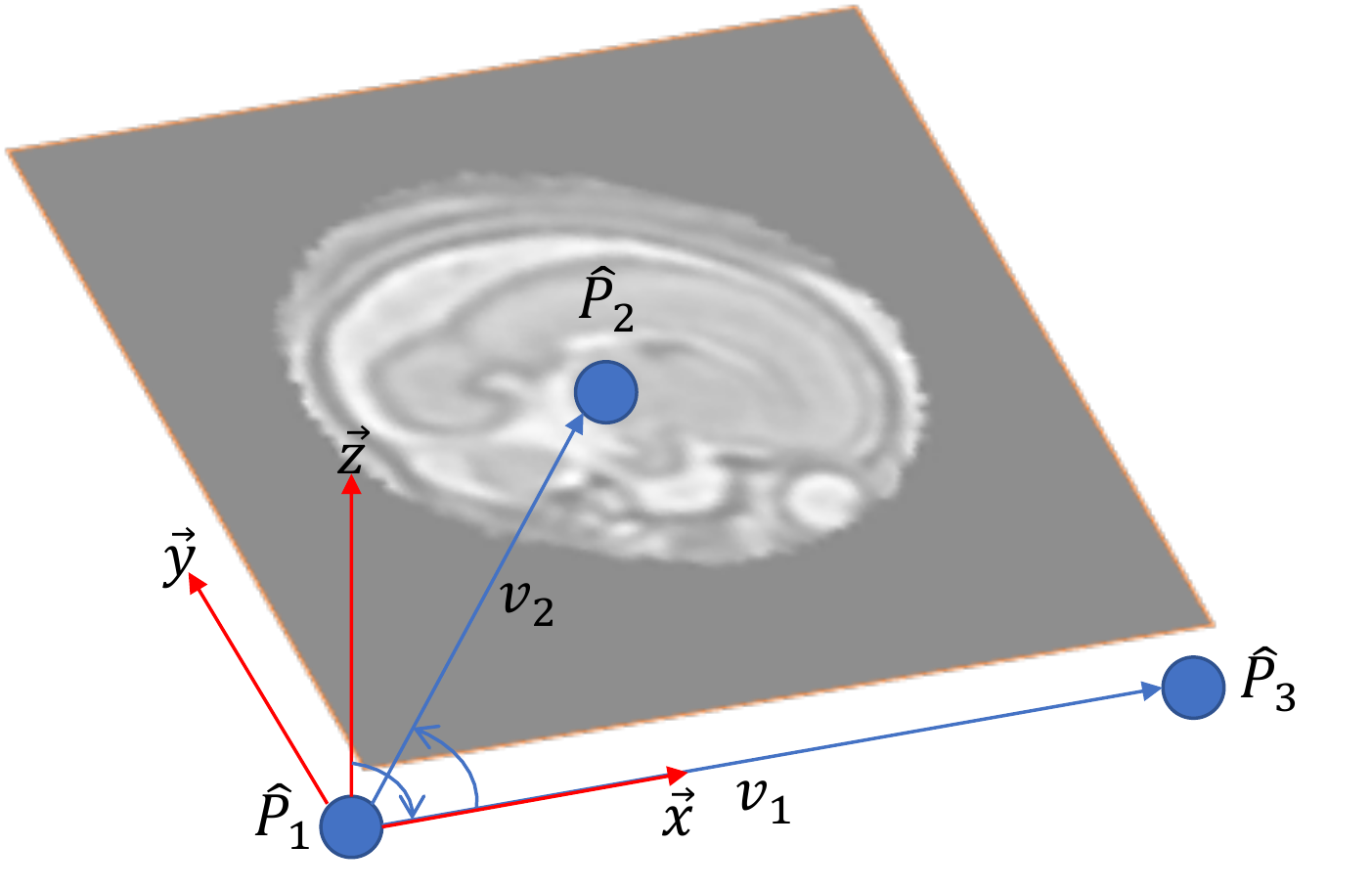}\label{fig:slice3d}}
  \\
  \subfloat[Base]{\includegraphics[height=1.2cm]{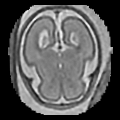}\label{fig:1a-base}} \hfill
  \subfloat[$T_x$]{\includegraphics[height=1.2cm]{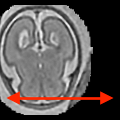}\label{fig:1b-tx}} \hfill
  \subfloat[$T_y$]{\includegraphics[height=1.2cm]{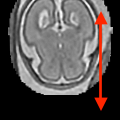}\label{fig:1c-ty}} \hfill
  \subfloat[$T_z$]{\includegraphics[height=1.2cm]{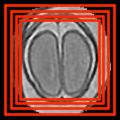}\label{fig:1d-tz}} \hfill
  \subfloat[$R_x$]{\includegraphics[height=1.2cm]{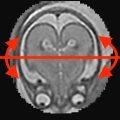}\label{fig:1e-rx}} \hfill
  \subfloat[$R_y$]{\includegraphics[height=1.2cm]{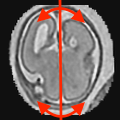}\label{fig:1f-ry}} \hfill
  \subfloat[$R_z$]{\includegraphics[height=1.2cm]{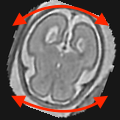}\label{fig:1g-rz}} \hfill
  \caption{
  (a-b) Visualization of extracting $\omega_i$ from $\Omega$. The identity plane lies flat on the $x$-$y$ axis, which is then rotated through an Euler Angle Iterator or a Fibonacci Point Iterator and then shifted up and down the normal $T_z$ (represented by red arrows) to account for out-of-plane transformations. Orange curved arrows represents rotations $R_x$, $R_y$ and $R_z$. (c) Anchor Point slice parameterization in 3D Space. (d-j) Transformations in 6 DoF.}
  \label{fig:samplingschema}
\end{figure}


\begin{figure}[htb]
  \centering
  \subfloat[Euler Angle]
  {\includegraphics[width=0.25\linewidth]{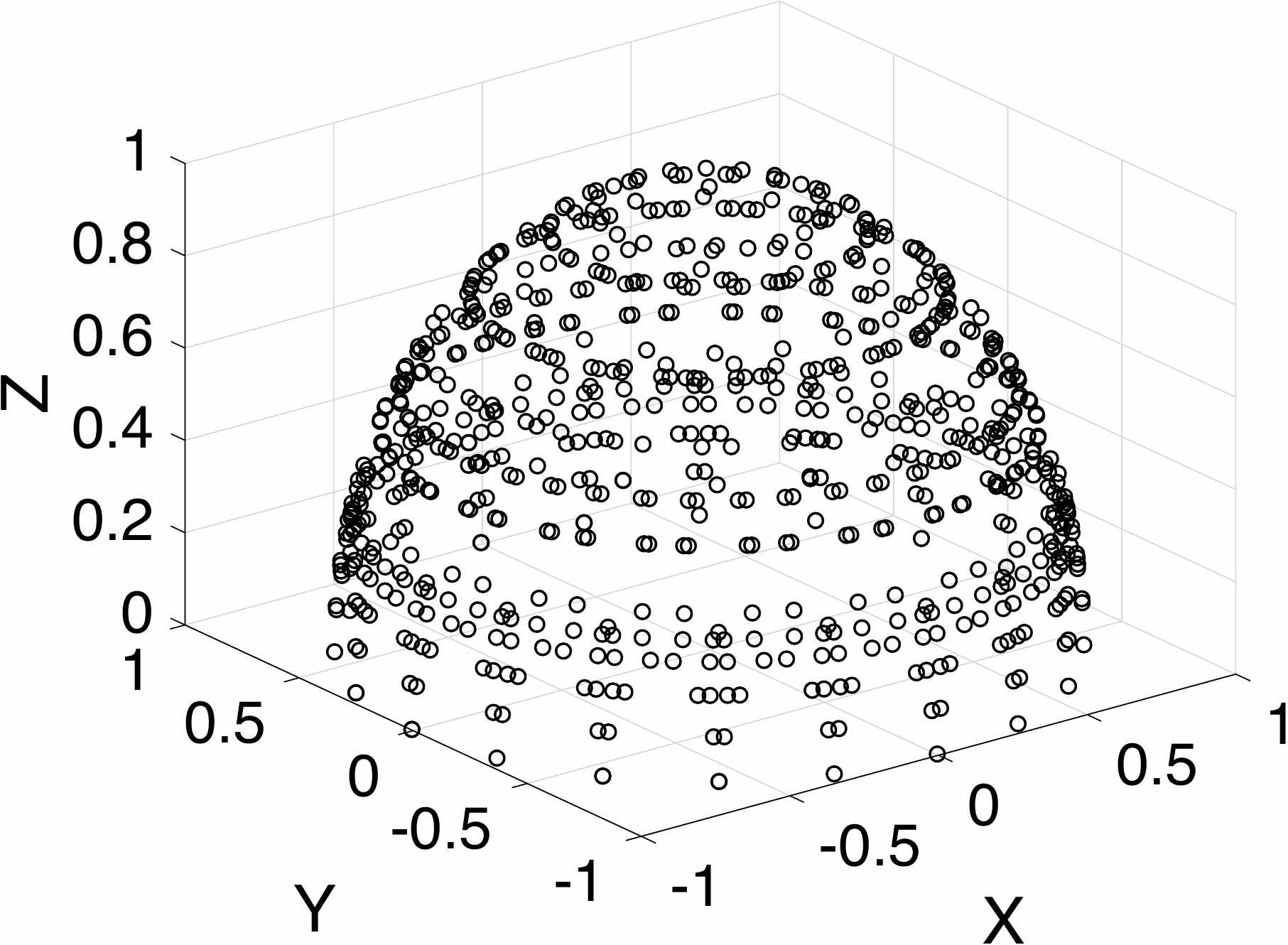}\label{fig:rp}} \hfill 
  \subfloat[Uniform Polar]
  {\includegraphics[width=0.25\linewidth]{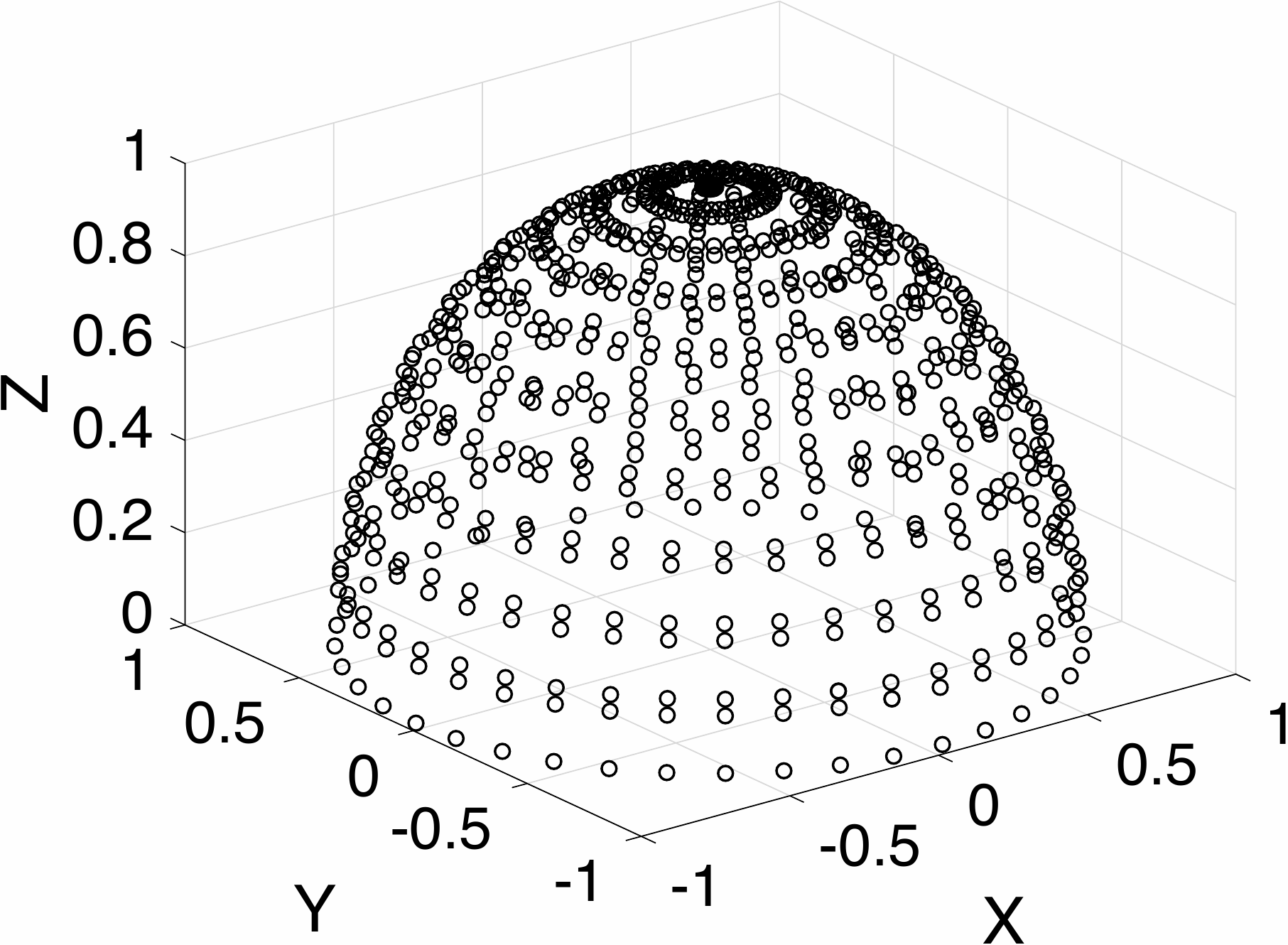}\label{fig:rp1}} \hfill
  \subfloat[Fibonacci]
  {\includegraphics[width=0.25\linewidth]{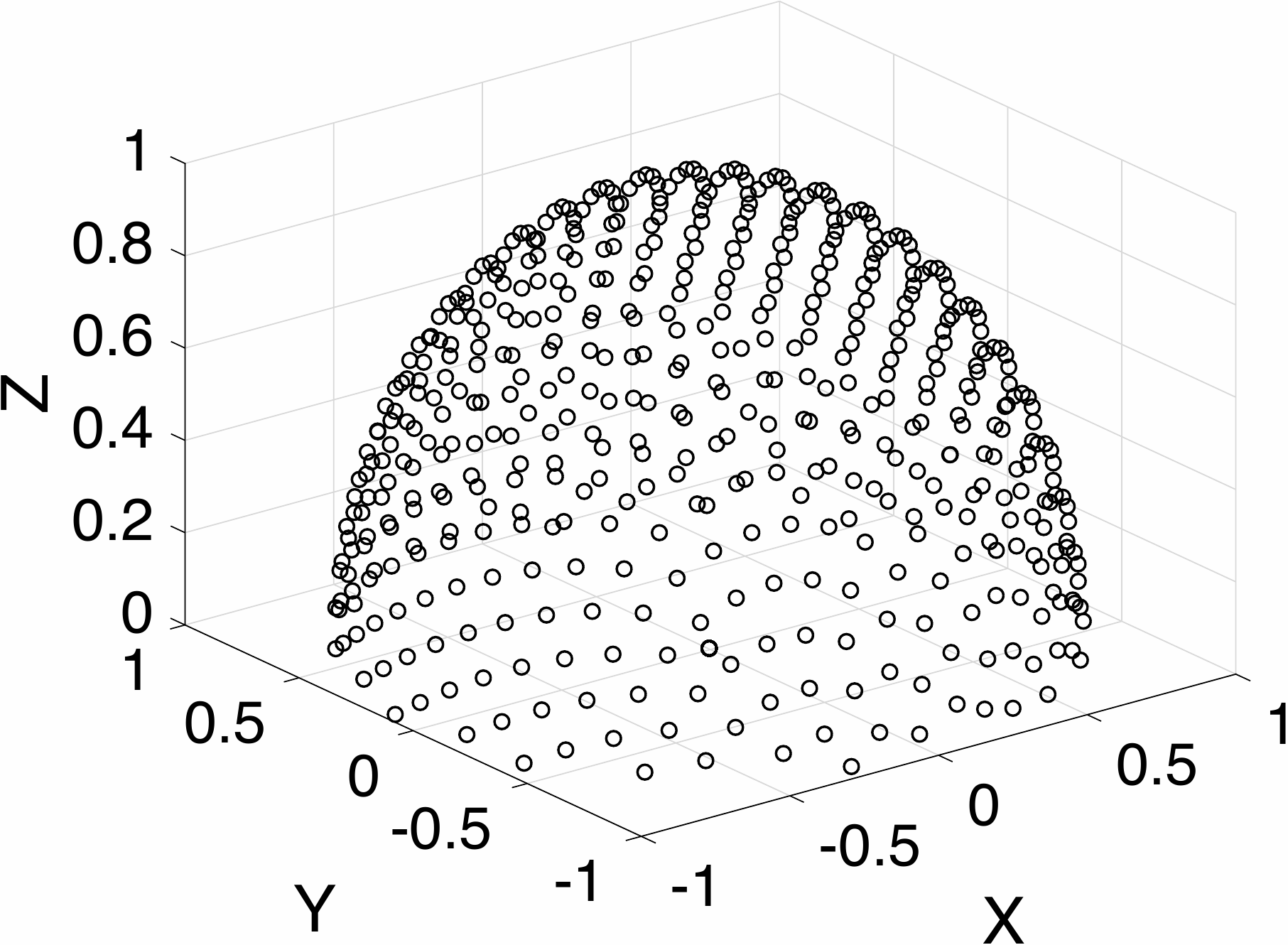}\label{fig:rp2}} \hfill
  \subfloat[Random]
  {\includegraphics[width=0.25\linewidth]{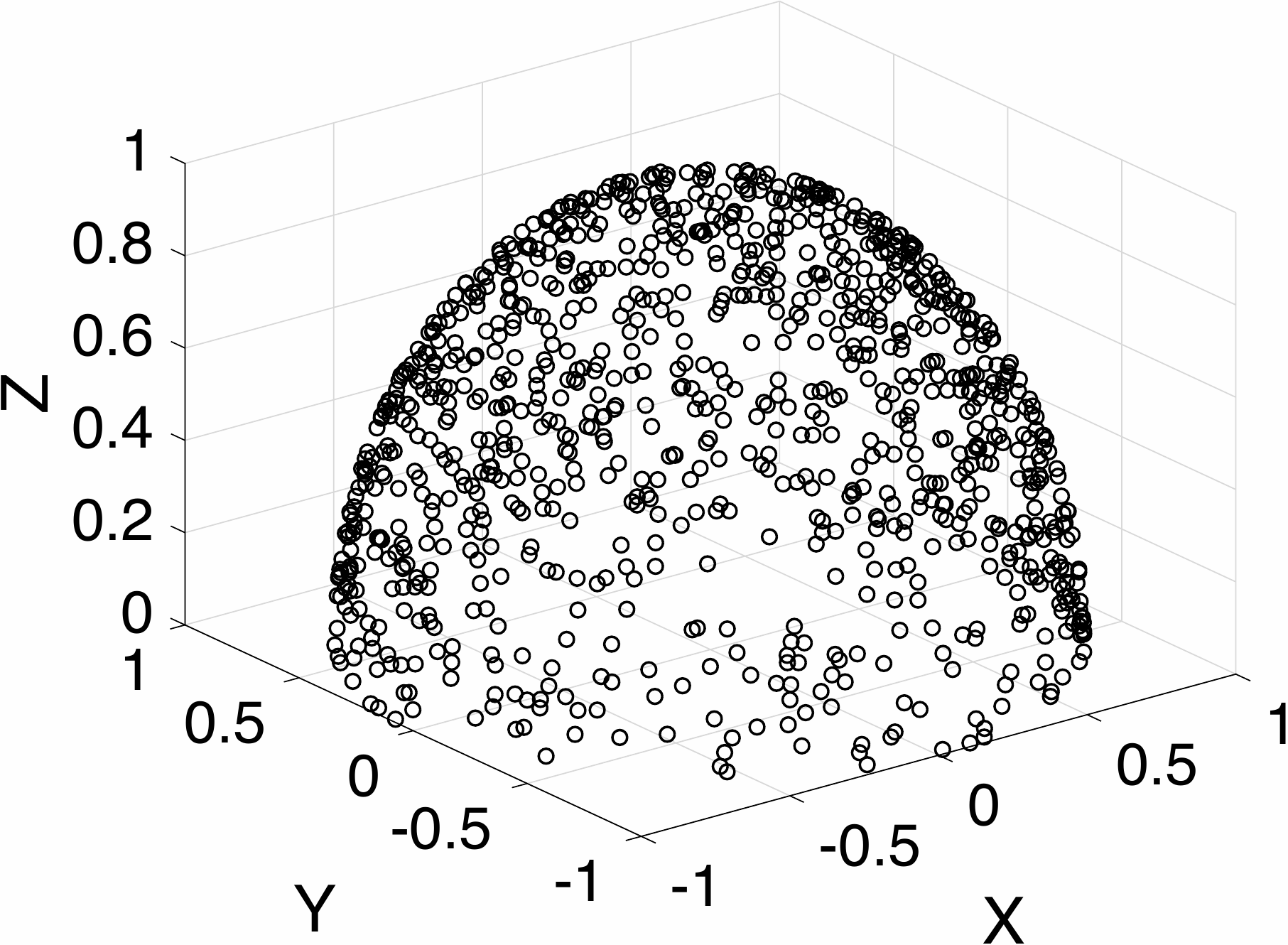}\label{fig:rp3}} \hfill
  \caption{Slice plane normals w.r.t the origin via different generation methods. The vector from the origin to each point cross through each slice origin. Note that it is not possible to visualize in plane rotation in this visualization}
  \label{fig:generation-methods}
\end{figure}




To create a comprehensive training and validation set, $\omega_i$  must cover a large number of transformation permutations in $\Omega$. We parameterize $\omega_i$ and $\Omega$ with the length $L$, such that the dimension of a slice $\omega_i$ matches the dimension of a face of the cubic volume $\Omega$. Every $\Omega$ encloses a brain atlas that is center-aligned (\emph{i.e.}, the origin is at the center of the brain) and which is isotropic and intensity normalized between 0 - 255. This is shown in Fig.~\ref{fig:samplingschema}a and \ref{fig:samplingschema}b. 

Since the transformation parameters are constrained to vary only $R_x$, $R_y$, $R_z$ and $T_z$, we rotate the sampling plane through each axis with multiple offsets, accounting for varying $T_z$. The identity plane, which lies flat on the $x$-$y$ axis, is initially rotated and then shifted up and down along the normal in the new orientation $T_z$ (represented by the red central axis in Fig.~\ref{fig:samplingschema}(a-b)). The ground truth transformation defines the transformation of the identity plane to its new and final location in 3D space.

A straightforward method is to iterate the rotation through Euler angles, where $R(x,y,z)$ and $[x,y,z] \in \mathcal{U}(-\pi/2,\pi/2)$. However, this does not give a balanced training slice distribution as shown in Fig.~\ref{fig:rp}. In this figure, each point represents the normal vector of the sampling planes from the origin (red arrow in Fig.~\ref{fig:samplingschema}a and \ref{fig:samplingschema}b). 
Another sampling approach uses polar co-ordinates, $P(\phi,\theta)$. Uniform sampling of polar co-ordinates causes an imbalance of samples, with a higher density of samples near the poles (Fig.~\ref{fig:rp1}). A better compromise is to use Fibonacci Sphere Sampling~\cite{González2009}, where each normal has roughly the same degree of separation as its neighbors (Fig.~\ref{fig:rp2}). The sampling normals are calculated using $P(\phi_i,\cos^{-1}(z_i))$, where $\phi_i = 2{\pi}i/\Phi$ and $z_i = 1 - (2i+1)/n$, $i \in {0,1,2,...,n-1}$. $\Phi = (\sqrt{5}+1)/2$ is the golden ratio and as such $\Phi^{-1}=\Phi-1$. We evaluate the impact of these training data sampling schemes in Sect.~\ref{sec:experiments}.

The rotation here is defined by the rotation required to transform a 3D vector $A$ onto a 3D vector $B$. Where $A$ is the starting normal, a unit vector in $z$ (\emph{i.e.}, $(0,0,1)$), and $B$ is the target sampling normal. However, this does not account for any in-plane rotation, $R_z$. The transformed sampling plane is further rotated around the Z-axis through a uniform distribution of angles, such that $R_z \in \mathcal{U}(0,\pi)$. 

For the validation set, slices are generated with the polar co-ordinates method using random normals and random in-plane rotation angles that are within the bounds of the training set. This is to simulate continuous motion, as test slices will not lie on a discrete training sampling interval, as shown in Fig.~\ref{fig:rp3}.

We constrain the shift along the normal, $T_z$, such that it encloses approximately the central 70\% of the volume and the range of $T_z$ is  $- 0.35L \leq T_z \leq 0.35L$. Edge cases are not beneficial to the training set, they contain little or no content. 
An edge slice can be ambiguous, its precise location cannot be determined without extra information even for a trained medical expert. Ambiguous samples of this nature can introduce adverse effects when training a CNN. Missing information from edge case slices can be recovered with intersecting orthogonal slices for eventual 3D reconstruction.



\subsection{Loss Functions and Ground Truth Labels}
The most commonly used loss functions for regression problems are Euclidean norms~\cite{krizhevsky2012imagenet,simonyan2014very}: $\left\| \hat{x} - x \right\| _n $. However, they may not be suitable when the regression target variables lie on a manifold that is non-Euclidean. For our proposed method, the slices are being transformed rigidly in 3D space, parameterization of each slice therefore lies within the bounds of the SE(3) Lie Group. This includes both; a rotation as well as a translation component. There are numerous ways of representing this transformation, such as for rotation; Euler angles, quaternions, rotation matrices, etc. 

To address the aforementioned challenge, Kendall et al.~\cite{Kendall_2015_ICCV} proposed the PoseNet loss as

\begin{equation}
	\textbf{Loss} = \left\| \hat{x} - x \right\| _2 + \beta \left\| \hat{q} - \frac{q}{\left\| q  \right\| } \right\| _2
\label{eqn:posenetloss}
\end{equation} 

which is used to regress the pose of a camera in 6 DoF. $x$ and $q$ are the predicted Cartesian translation and quaternion rotation parameters, whilst $\hat{x}$ and $\hat{q}$ are the respective ground truth values. This loss function combines the Euclidean distance of translational loss with a weighted Euclidean distance of the rotation loss. $\beta$ is a tuning parameter that is used to determine the contribution between the losses, by normalizing numbers with different scales. Quaternions can represent a rotation by using four numbers between $+1$ and $-1$, however Cartesian 3 DoF co-ordinates span from $-\infty$ and $+\infty$. This causes an imbalance for combined optimization and requires manual correction through $\beta$.

Xu et al.~\cite{Xu:2016:MRD:3024223.3024275} have proposed a framework based on separating loss functions that has an advantage of alleviating over-fitting. The fully connected layer of the network is split into several branches, where each branch is terminated with a separate loss function. Instead of manually tuning the contribution of discrete components in a combined loss function, the network incorporates the tuning parameter within the connection weights. As a result, the network is able to learn from multiple representations of the ground truth labels, for instance, Euler-Cartesian parameters (3 for rotation and 3 for translation) and Quaternion-Cartesian parameters (4 for rotation and 3 for translation).

We introduce a novel labelling system where the rotation and translation components of the labels are combined together. Any three non co-linear points in a 3D Euclidean space form a plane, while their order defines the orientation. We therefore call them {\em Anchor Points} (see Fig.~\ref{fig:samplingschema}(a-c)). 

Three Anchor Points can be defined anywhere on an $L \times L$ 2D slice $\omega_i$, as long as they are not identical or co-linear and the relative in-pane locations are consistent throughout all slices in the data set. For simplicity, we defined $P_1$ to be the bottom-left corner $(-L,-L,0)$, $P_2$ to be the origin $(0,0,0)$ and $P_3$ the bottom right corner $(L,-L,0)$ on the Identity plane. Fig.~\ref{fig:samplingschema} shows Anchor Points being marked on multiple slices, and Fig.~\ref{fig:slice3d} shows the Anchor Point on one particular slice. Anchor Points and the identity sampling plane are both transformed to their destined location using the same transform parameter set. Consequently, Anchor Point labels consists of 9 parameters: ($P_1(x,y,z)$, $P_2(x,y,z)$ and $P_3(x,y,z)$). As each point is Cartesian, the optimization is balanced and it can be calculated with the standard L2-norm loss function. Incorporating the multi-loss framework~\cite{Xu:2016:MRD:3024223.3024275}, the losses for $P_1$, $P_2$ and $P_3$ are calculated independently. The combined loss for SVRnet can therefore be written as: 

\begin{equation}
    { \textbf{Loss}} = \alpha \left\| \hat { { P }_{ 1 } } -{ P }_{ 1 } \right\| _{ 2 } +  \beta\left\| \hat { { P }_{ 2 } } -{ P }_{ 2 } \right\| _{ 2 }  +  \gamma\left\| \hat { { P }_{ 3 } } -{ P }_{ 3 } \right\| _{ 2 } 
\label{eqn:svrnetloss}
\end{equation} 

\subsection{Network Architecture and Uncertainties}\label{sec:MNU}
\label{sec:nets}
Towards making appropriate network architecture choices for SVRnet, we explore several state-of-the-art networks:  CaffeNet~\cite{jia2014caffe}, GoogLeNet~\cite{43022}, Inception~\cite{journals/corr/SzegedyIV16}, NIN~\cite{DBLP:journals/corr/LinCY13}, ResNet~\cite{DBLP:journals/corr/HeZRS15} and VGGNet~\cite{DBLP:journals/corr/SimonyanZ14a}. SVRnet takes $\omega_i$ as inputs whilst computing the loss in Eq.~\ref{eqn:svrnetloss} of various labelling methods. In Sect.~\ref{sec:experiments} we evaluate each architecture on the regression performance of the previously proposed Anchor Point labels.


A common strategy during the training of such large, state-of-the-art networks involve the use of dropout~\cite{srivastava2014dropout}. This entails muting components of the true signal, provided to individual neurons. The technique essentially provides a form of model averaging. Dropout constitutes a well-understood regularization technique to reduce over-fitting. As a result of dropout, neurons have the ability to produce different outputs upon successive activations. During inference, dropout is usually disabled such that network consistency is not undermined. Regression networks are therefore commonly deterministic models at inference time and do not allow for the modelling of uncertainty. Implementing fully probabilistic models that account for uncertainty in both (1) the data and (2) the model parameters (Aleatoric, Epistemic uncertainties respectively~\cite{DBLP:journals/corr/KendallG17}) may introduce high computational cost~\cite{COOPER1990393}.

Gal et al.~\cite{pmlr-v48-gal16} recently showed that dropout layers in Neural Networks can be interpreted as a Bayesian approximation to probabilistic models, and can be implemented by applying a dropout before every weight layer. This is shown to be mathematically equivalent, as it approximately integrates over the models’ weights.
The authors further show in~\cite{Gal2016Bayesian} that, for the same input, performing multiple predictions during test time with dropout and taking a mean of the predictions improves result for CNN based networks. 
This process of performing multiple predictions from the same input by using dropout layers is called \emph{Monte Carlo Dropout sampling}, which also provides model uncertainty for the given input data.

Using this technique, our experimental work in Section~\ref{ref:recon} focuses on taking epistemic uncertainty into consideration in order to gauge alignment prediction confidence in real-world test cases. Slice alignment requires high precision, and we investigate the idea that a measure of prediction confidence is important to aid reconstruction quality. 

\lastrev{Network prediction confidence is also used as a metric to screen out corrupted slices, \emph{i.e.}, regions of the image with signal dropouts or intensity bleeding from amniotic fluid. Our data predominately show's signal dropout artefacts but network prediction confidence can be used to reject any kind of image corruption.} An obscured image slice would result in a prediction with low confidence. Slices with low confidence can be discarded and not further used for subsequent 3D reconstruction. 





\subsection{Metrics on Non-Euclidean Manifolds}\label{sec:metrics}
Computing the mean prediction of the Monte Carlo Dropout samples requires an accurate method of averaging the network output, which, in our case, is a rigid transformation.
Rigid transformations do not lie on the Euclidean manifold, but constitutes of a smooth manifold where an intrinsic mean and corresponding variance can be computed~\cite{pennec2006intrinsic}. This is more commonly regarded as the Special Euclidean group SE(3).

Considering $N$ rigid transformations predicted by the network: $\{x_i\}; \quad i = 1,...,N$, we compute a Riemannian center of mass that minimizes geodesic distances between all points:
\begin{equation}
    m = \underset{y \in \mathcal{M}}{\operatorname{argmin}} \, \mathbf{E}\left[ \textrm{dist}(x,y)^2 \right] 
\label{}
\end{equation}
where $\mathcal{M}$ is a Riemannian manifold and $\text{dist}(x,y)$ defines a geodesic distance between two points $x$ and $y$ on this manifold. 
A Gauss-Newton iterative algorithm on rigid transformations is used to compute such a mean, $m$, from the available data points $x_i$ by using a left-invariant metric \cite{pennec2006statistical} 
\begin{equation}
    m_{t+1} = m_t \circ \exp_{Id} \left( \frac{1}{n}\Sigma{\log_{Id}\left( m^{-1}_t \circ x_i \right)} \right) 
\label{}
\end{equation}
where $\exp_{Id}$ and $\log_{Id}$ are the exponential and logarithmic mappings from identity as defined by the left-invariant Riemannian metric, and $\circ$ is the group composition operator.
Once the mean is computed, the corresponding variance is straightforward to compute as $\sigma = \mathbf{E}[(x-m)^2]$, where the logarithmic operator defines $x-m$ as $\log_m(x)$.
\cite{pennec2006statistical} provides a detailed overview for these notions and algorithms.

\subsection{Transformation Recovery}
We transform the predicted Anchor Point positions, Euler-Cartesian and Quaternion-Cartesian parameters to a rotation matrix and Cartesian offset. We then transform the corresponding slice to its inferred location in 3D space. However, the network may introduce a prediction error, which will likely cause the Anchor Points to deviate from a perfect isosceles triangle formation. To overcome this problem, we assume:

\begin{itemize} 
  \item $P_2$ defines the Cartesian offset of the slice, \emph{i.e.}, the center point of the slice in world space.
  \item The vector joining points $P_1$ and $P_3$ aligns with the bottom edge of the slice (this defines the in-plane rotation).
  \item The Anchor Points together define the plane, which contains the slice. It is used to calculate the rotation matrix to reorient the slice (see Fig.~\ref{fig:slice3d}).
\end{itemize}


The Cartesian offset, $t$ equals $P_2$. For the orientation. We calculate three orthogonal vectors that defines the new co-ordinate system. Concatenating these vectors gives the rotation matrix, which transforms an identity plane to the newly predicted orientation. $\vec{v_1} = P_3 - P_1$ and $\vec{v_2} = P_2 - P_1$ (represented by blue arrows in Fig.~\ref{fig:slice3d}).
$\vec{v_1} \times \vec{v_2}$ gives the normal of the plane $\vec{n_1}$, this defines the new Z-axis. $\vec{n_1} \times \vec{v_1}$ gives the new Y-axis $\vec{n_2}$, and $\vec{v_1}$ itself defines the new X-axis. Finally, $\vec{v_2}$, $\vec{n_2}$ and $\vec{n_1}$ are concatenated together to get the rotation matrix $R$. 

It is important to note that Anchor Points are unable to coincide with one another, by definition. Each anchor point is regressed using an independent fully connected network layer and, as such, two (or more) Anchor Points may only coincide if their fully connected layer weights would be identical. Layer weights are randomly initialised and will deviate from each other during training due to the pre-defined, non-identical, training sample target locations.
It is possible that two Anchor Points are predicted in close proximity in rare error cases. Cases of this type are identified by checking the constraint that Anchor Points must adhere to a minimum distance from one another.


\subsection{Slice to Volume Reconstruction}
We use a modified version of \cite{kainz2015fast} to perform Slice to Volume Reconstruction on the individual $\omega_i$. Instead of using a pre-existing volume as the initial registration target we create an initial registration target volume from all $\omega_i$ and their corresponding predicted $\hat{T_i}$. 

For validation cases, $\omega_i$ that are utilized for inference are also used for reconstruction. This allows us to verify whether or not the original volume $\Omega$ can be recovered from $\omega_i$ and $\hat{T_i}$, since $\omega_i$ were extracted from $\Omega$ for the training and validation data sets. 
In Sect.~\ref{sec:experiments} we employ image quality metrics to assess the prediction performance by examining the original and reconstructed volumes.

For test cases, intensity rescaled $\omega_i$ or original $\omega_i$ are used for slice-to-volume reconstruction. The intensity rescaled image is only required for SVRnet prediction, during which, the top 1\% and bottom 1\% of the intensity distribution is also pruned to ensure the distribution is not skewed by outliers. Quantizing from 16bit to 8bit reduces the Signal to Noise Power Ratio (SNR) from 96dB to 48dB. 48dB SNR (calculated by $\text{SNR} = 20\log_{10}(2^{(16-8)}) = 48.2\text{dB}$) is still sufficient for accurate identification of important image features through SVRnet.



\section{Implementation}\label{sec:impl}

All network architectures were implemented using the Caffe~\cite{jia2014caffe} library on an Intel i7 6700K CPU with Nvidia Titan X Pascal GPU. All fetal subject data were provided by the iFIND project \cite{ifind}. Scans were performed on a Philips Achieva 1.5T at St Thomas Hospital, London, with the mother lying 20$^\circ$ tilt on the left side to avoid pressure on the inferior vena cava or on her back depending on her comfort. For each subject, multiple Single Shot Fast Spin Echo (SSFSE) images were acquired with in-plane resolution of 1.25 $\times$ 1.25 mm and slice thickness of 2.50 mm. The gestational age of the fetuses at the time of scans was between 21 to 27 weeks (mean=24, stdev=1). 
All slices are of size $120 \times 120$ (\emph{i.e.}, $L=120$). Six separate data sets were generated to cover combinations of slice generation and label representations. Three data sets were generated via the Euler angle iteration, with the remaining three generated using the Fibonacci Sphere Sampling method. Of each slice generation method, there is a data set for Euler-Cartesian labels, Quaternion-Cartesian labels and Anchor Point labels.

For the Euler generation method, angles are iterated through $18^{\circ}$ intervals from $-90^{\circ}$ to $+90^{\circ}$, which gives 10 sampling intervals for each axis. 40 slices are taken in the $T_z$ axis, such that $-40 \leq T_z \leq +40$ in 2 mm intervals. This samples approximately the middle 66\% of the volume. In total, the Euler iteration method generates 2.24M images for this training set.

For the Fibonacci Sphere Sampling method, 300 sampling normals were chosen. This gives approximately $8^{\circ}$ separation between every normal. An additional 10 images, between $0^{\circ}$ and $180^{\circ}$ with $18^{\circ}$ interval, were generated at each normal to account for in-plane rotation. Slices are also taken between $-40$ and $+40$ in $T_z$ with 4 mm interval. This generates in total 3.36M images for this training set. 

The validation slices are generated in similar fashion, except that they are sampled at random intervals within the bounds of the training set 
to model random subject motion.

We implemented~\cite{Miolane2013} in Python  for mean, variance and geodesic distance computations on SE(3) groups of rigid transformations.

\section{Experimentation and Evaluation}
\label{sec:experiments}

\subsection{Evaluation Metrics}\label{sec:eval}
A na{\"i}ve progress check involves monitoring training and validation losses to ensure that the network is generalizing. To examine a slice in detail, we present the network with a 2D image slice $\omega_i$, as extracted from $\Omega$. Using the parameters obtained from the network during inference, we extract a new slice from the same $\Omega$ and compare it to slice $\omega_i$. Comparison is performed via several standard image similarity metrics, outlined in this section.

We utilize standard image processing metrics; Cross Correlation (CC), Peak Signal-to-Noise Ratio (PSNR), Mean Squared Error (MSE) and Structural Similarity (SSIM). As these metrics do not definitively assess slice location in 3D space, we included Euclidean Distance Error (average distance between the predicted and ground truth Anchor Points) as well as Geodesic Distance Error (a unitless metric between two SE(3) poses).

CC measures the similarity between two series (or images) as a function of the displacement of one relative to the other. This searches for features that are similar in both images by examining the pixel intensities 
\begin{equation}
	\hbox{CC}(\hat{f},\hat{g}) = \sum_{i}^{m-1}\sum_{j}^{n-1} \hat{f}(i,j)\hat{g}(i,j) 
\label{eqn:ncc1}
\end{equation}
where $N$ is the number of pixels in the image,
\begin{equation}
	\hat{f} = \frac{f - \bar{f}}{\sqrt{\sum(f - \bar{f})^2}} 
	\quad \textrm{and} \quad 
	\hat{g} = \frac{g - \bar{g}}{\sqrt{\sum(g - \bar{g})^2}}
\label{eqn:ncc2}
\end{equation}

PSNR (based on the MSE metric) is a ratio between the maximum possible power of a signal and the power of corrupting noise that affects the fidelity of its representation. This is the delta-error that is estimated by the network during regression and defined as 
\begin{equation}
	\hbox{PSNR} = 10 \cdot \log_{10}(\frac{MAX_I^2}{MSE})
\label{eqn:psnr}
\end{equation}
where $MAX_I$ is the maximum possible intensity (pixel value) of the image and
\begin{equation}
	\hbox{MSE}(f,g) = \frac{1}{mn} \sum_{i}^{m-1}\sum_{j}^{n-1} {(f(i,j) - g(i,j))}^{2}
\label{eqn:mse}
\end{equation}

SSIM attempts to improve upon PSNR and MSE, and uses a combination luminance, contrast and structure to assess the image quality. Metrics such as MSE can give a wide variety of degraded quality images with drastically different perceptual quality, which is an undesirable trait. It is defined as 
\begin{equation}
	\hbox{SSIM}(f,g) = \frac{(2\mu_f\mu_g + c_1)(2\sigma_{fg} + c_2)}{(\mu_f^2 + \mu_g^2 + c_1)(\sigma_f^2 + \sigma_g^2 + c_2)} 
\label{eqn:ssim}
\end{equation}
where $\mu_f$ and $\mu_g$ are the average pixel intensities of images $f$ and $g$ respectively; $\sigma_f^2$ and $\sigma_g^2$ are the variances of $f$ and $g$ respectively; $\sigma_{fg}$ is the co-variance of $f$ and $g$; $c_1=(k_1L)^2$ and $c_2=(k_2L)^2$ are two variables that stabilizes the division with weak denominator (\emph{i.e.}, cases where $\mu_f^2 + \mu_g^2 \rightarrow 0$ or $\sigma_f^2 + \sigma_g^2 \rightarrow 0$); $L$ is the maximum possible intensity of the image; with $k_1=0.01$ and $k_2=0.03$ are default constants. 




The average Euclidean Distance error, which is defined as 
\begin{equation}
\text{ E.D. } = \frac{1}{3} ( \left\| \hat { { P }_{ 1 } } -{ P }_{ 1 } \right\| _{ 2 } +  \left\| \hat { { P }_{ 2 } } -{ P }_{ 2 } \right\| _{ 2 }  +  \left\| \hat { { P }_{ 3 } } -{ P }_{ 3 } \right\| _{ 2 } ) 
\label{}
\end{equation} 

is the average Euclidean Distance of all three Anchor Points, and provides an error estimation in mm.

The geodesic distance on the prediction manifold provides an intrinsic distance measure of how far the predicted rigid transformation is from the ground truth transformation. If $x$ is a ground truth rigid transformation and $y$ is a predicted rigid transformation, a left-invariant geodesic distance with a metric at the tangent space of $x$ can be computed as \cite{pennec2006statistical}:
\begin{equation}
    \text{G.D.} = \text{dist}(x, y) = \left\lVert \log_x(y) \right\rVert_x 
\end{equation}
where $\log$ uses the same notion as described in Sect.~\ref{sec:MNU}. 


\begin{figure*}[!htbp]
  \centering
  \subfloat{\includegraphics[width=0.3\linewidth]{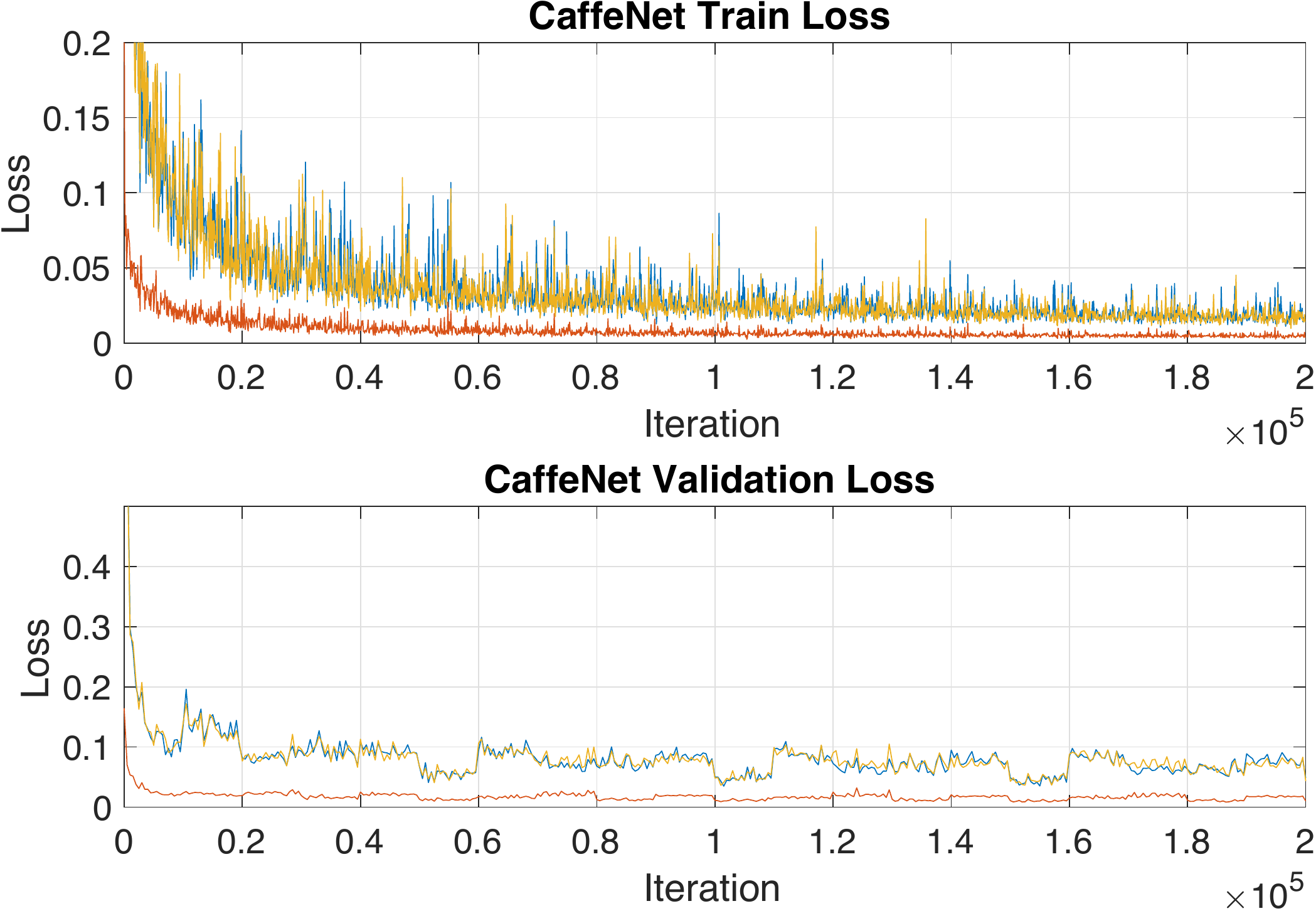}}      \hfill
  \subfloat{\includegraphics[width=0.3\linewidth]{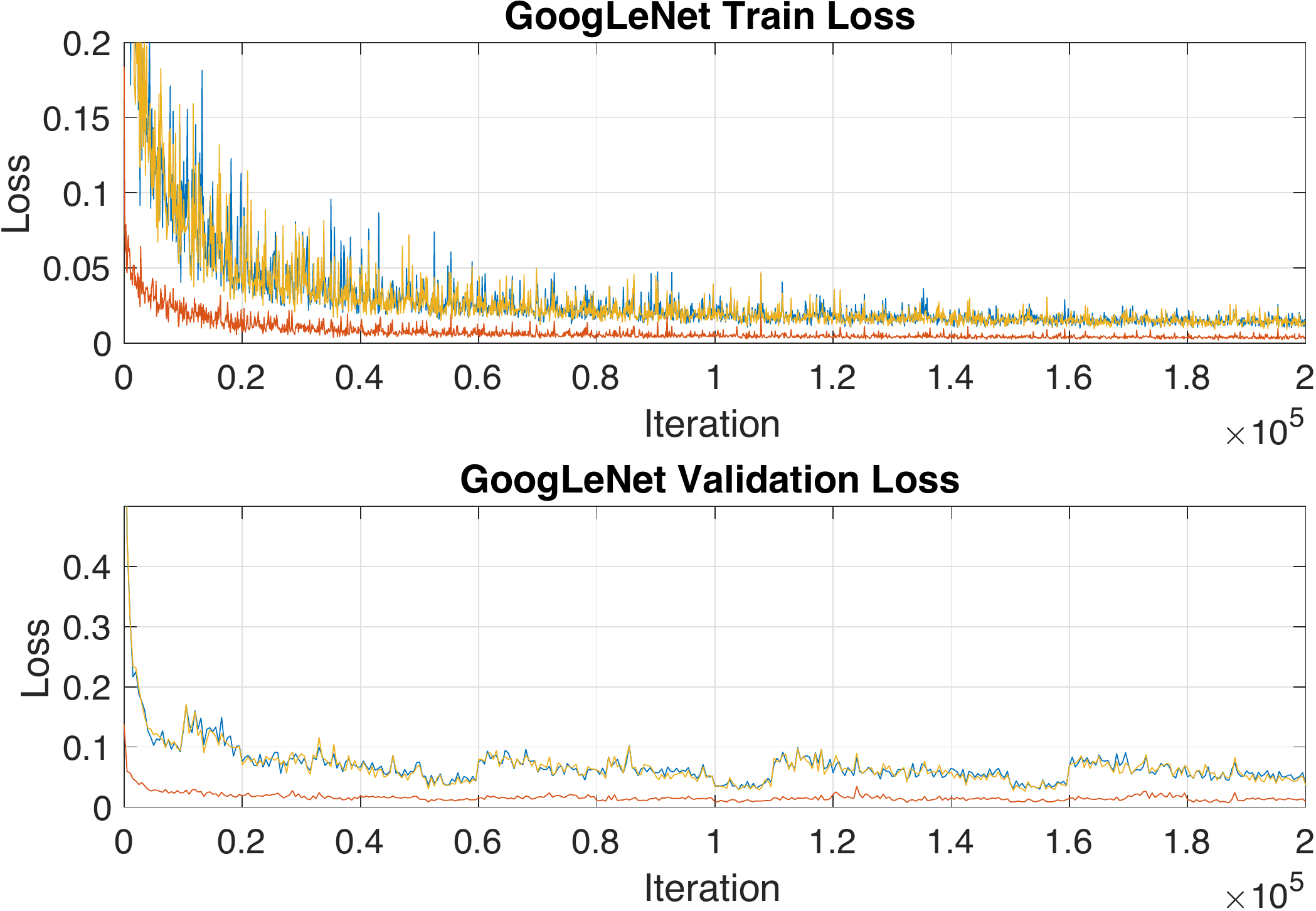}}     \hfill
  \subfloat{\includegraphics[width=0.3\linewidth]{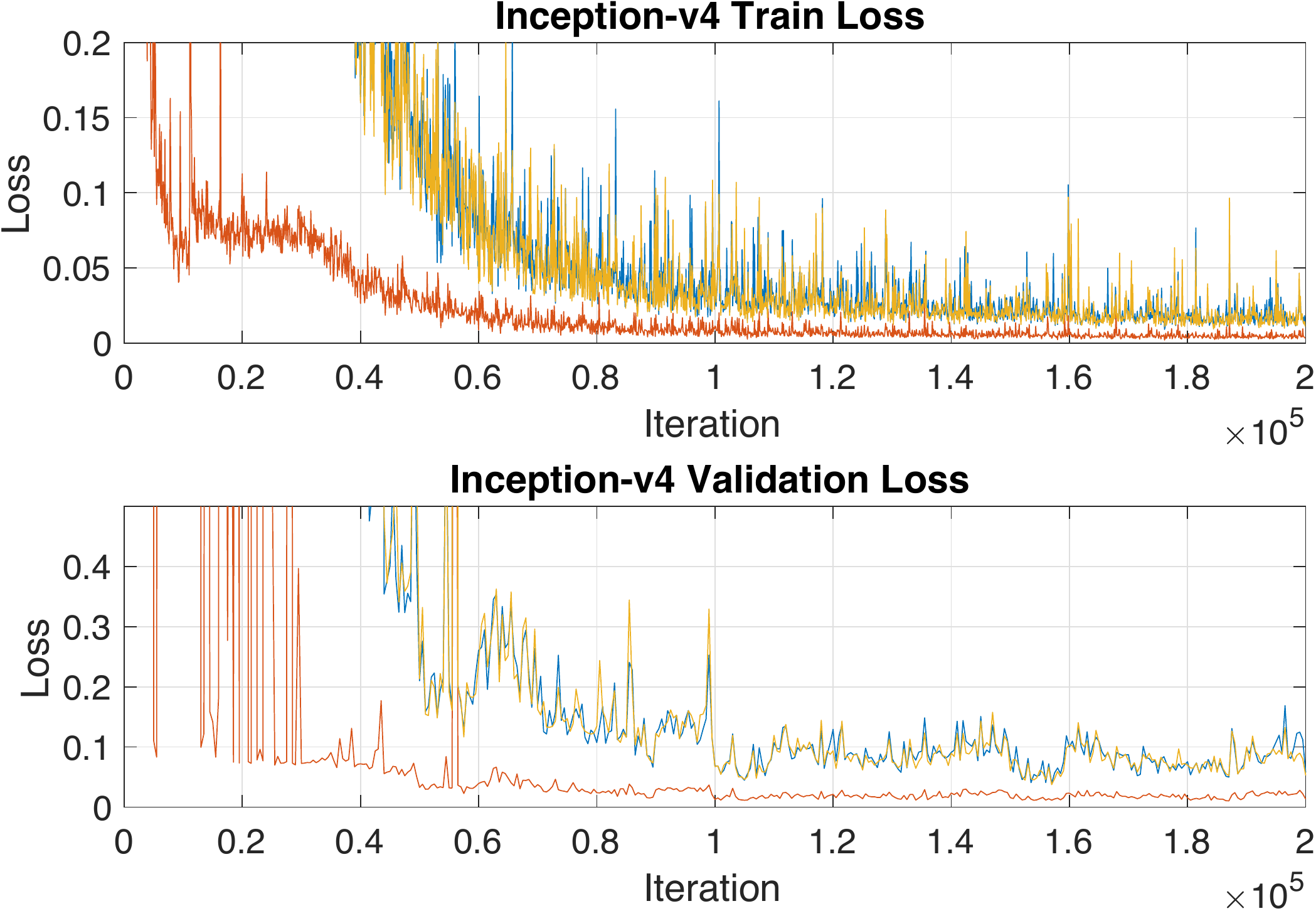}}  \hfill \\
  \subfloat{\includegraphics[width=0.3\linewidth]{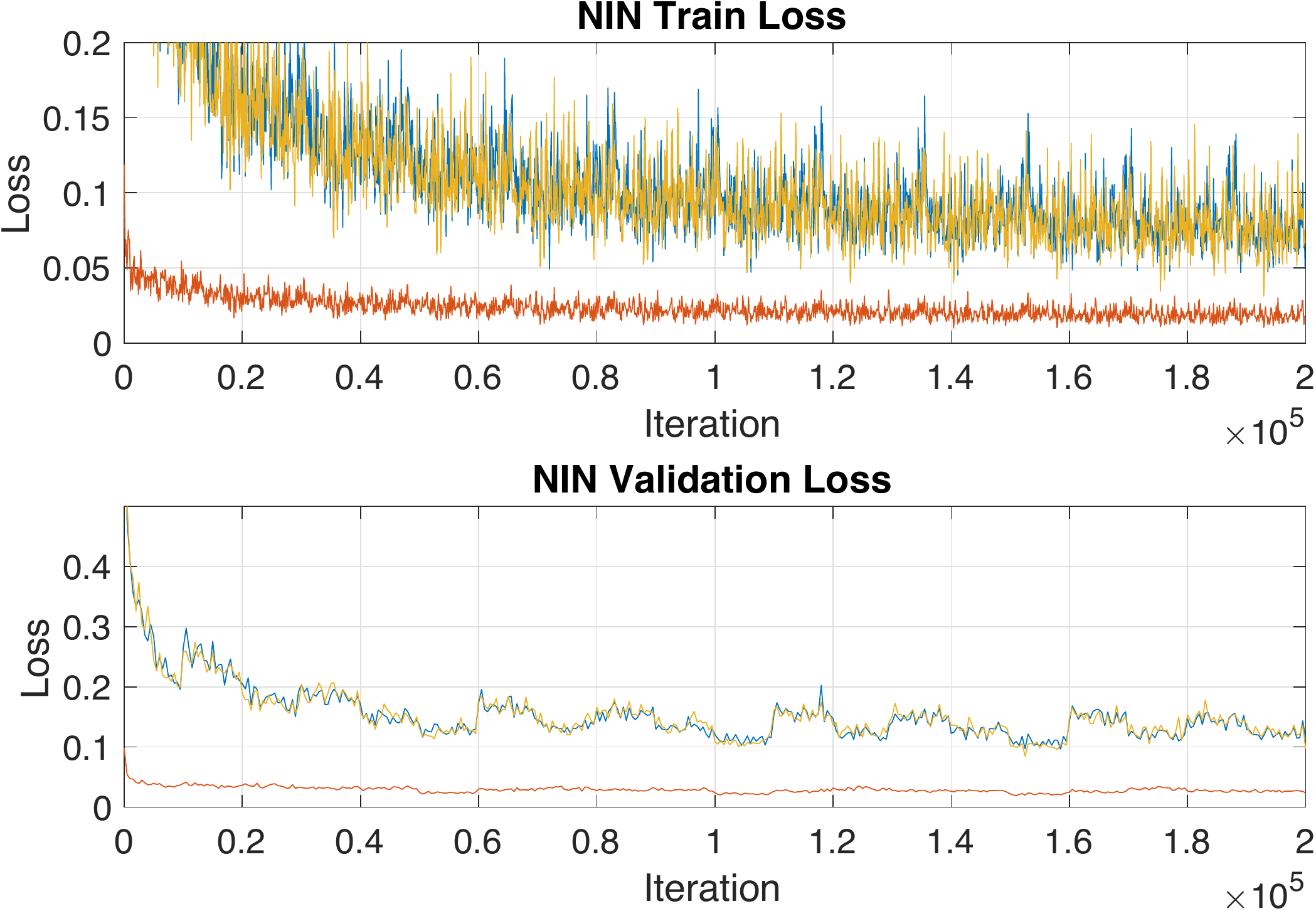}}           \hfill
  \subfloat{\includegraphics[width=0.3\linewidth]{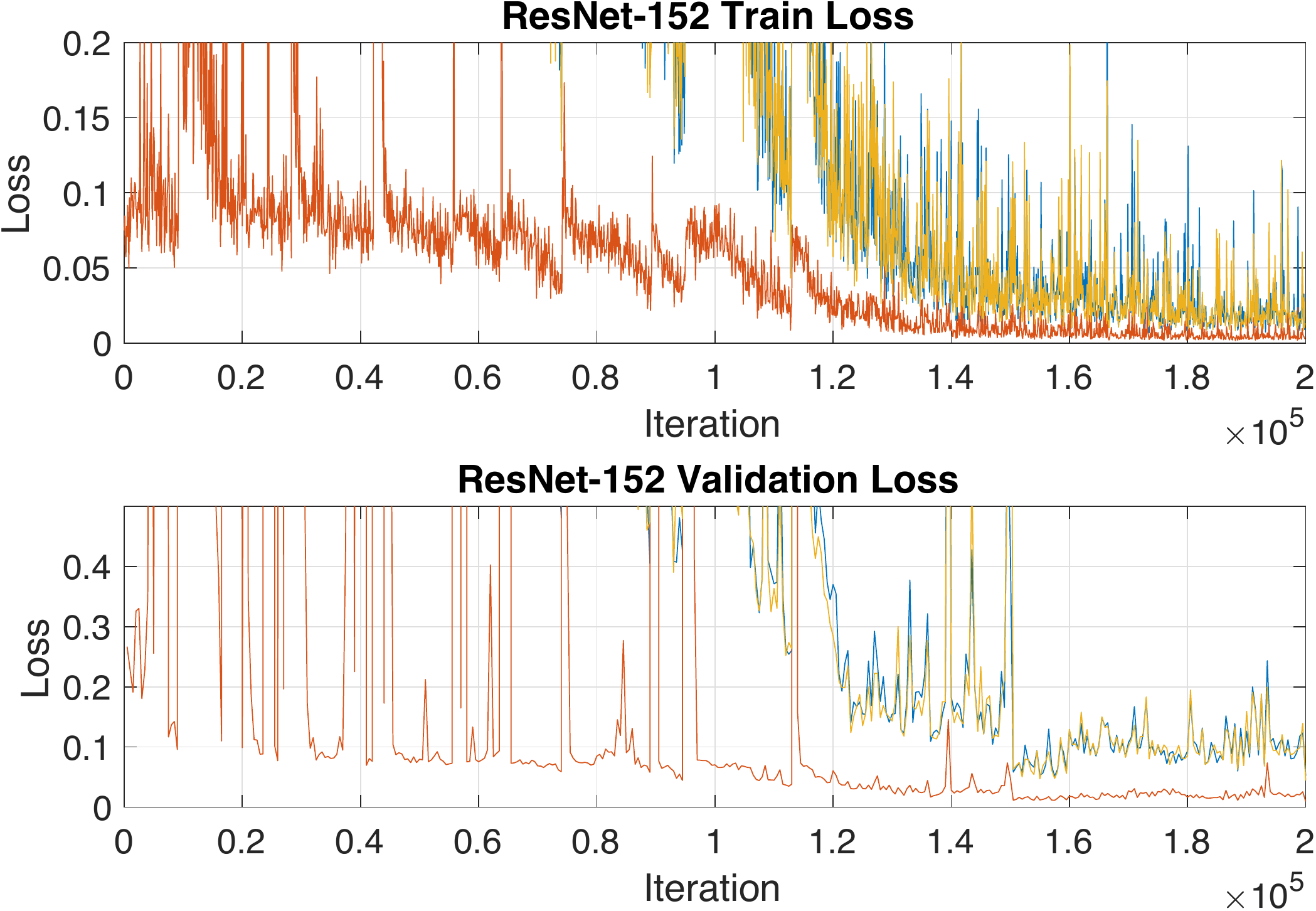}}    \hfill
  \subfloat{\includegraphics[width=0.3\linewidth]{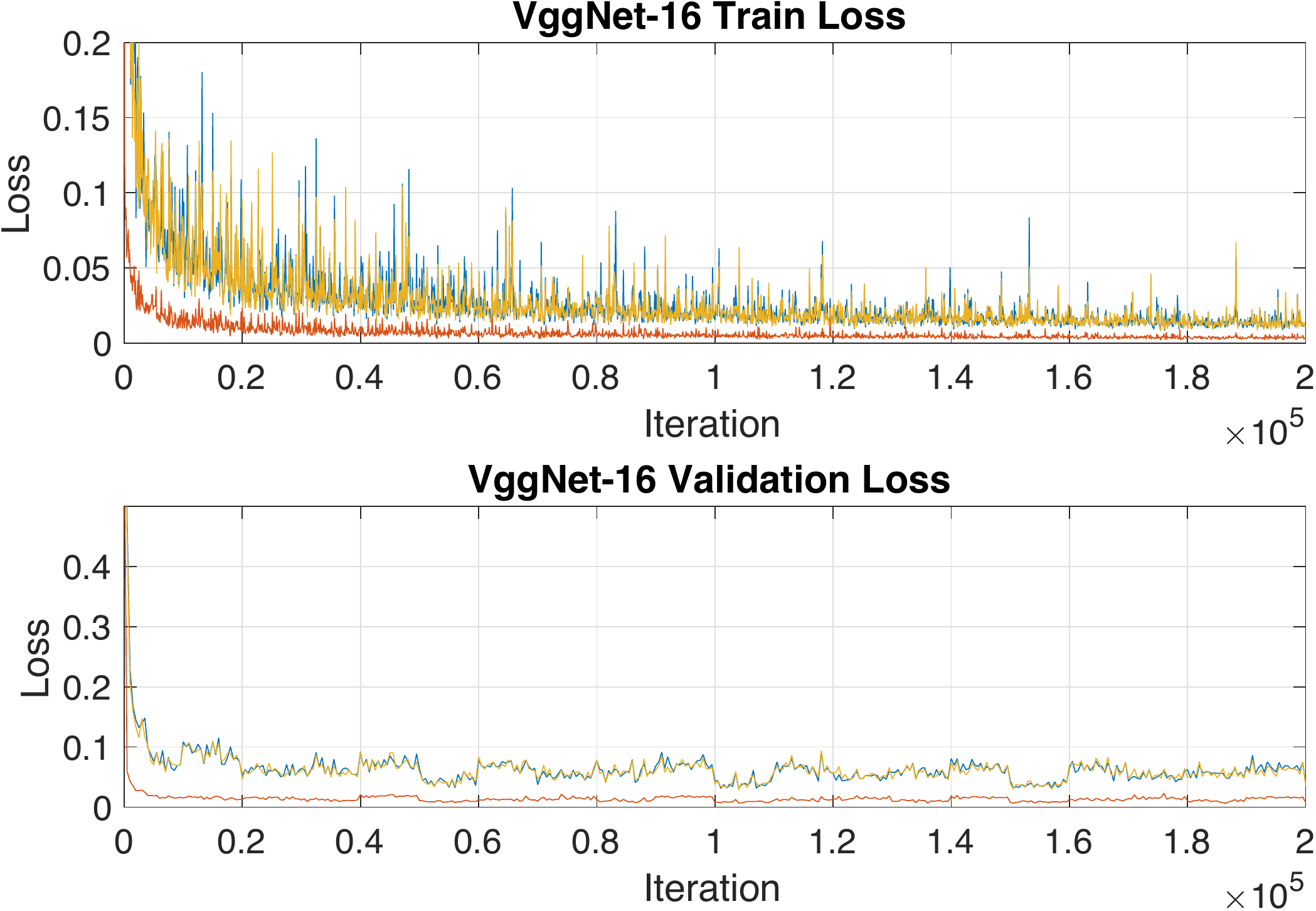}}     \hfill
  \caption{Loss graphs of Training (rows 1, 3) and validation (rows 2, 4) of the six network architectures used for Anchor Point predictions.
  }
  \label{fig:train-test-loss}
\end{figure*}

\begin{table*}[htb]
\centering
\caption{Table showing the mean error and standard deviation of each evaluation metric}
\label{tab:network-performance}
\begin{tabular}{@{}ccccccccccccc@{}}
\toprule
             & \multicolumn{2}{c}{CC} & \multicolumn{2}{c}{MSE}  & \multicolumn{2}{c}{PSNR} & \multicolumn{2}{c}{SSIM} & \multicolumn{2}{c}{Avg E.D. Error (mm)} & \multicolumn{2}{c}{~~~Avg G.D. Error~~~} \\ 
\cmidrule(l){2-13} 
             & $\bar{\mu}$     & $\sigma$      & $\bar{\mu}$ & $\sigma$    & $\bar{\mu}$  & $\sigma$   & $\bar{\mu}$  & $\sigma$   & $\bar{\mu}$      & $\sigma$        & $\bar{\mu}$      & $\sigma$        \\ 
\midrule
CaffeNet     & 0.8137          & 0.1014        & 1085.1          & 584.8715    & 18.4963      & 2.5625     & 0.5327       & 0.1705     & 8.3558           & 4.4449      & 20.1427          & 19.1897         \\
GoogLeNet    & 0.8378          & 0.1014        & 935.04          & 576.4873    & \textbf{19.3564} & 2.9482 & \textbf{0.5845} & 0.1775  & \textbf{7.1498}  & 4.8049      & 15.7504          & 15.8088         \\
~~Inception v4~~ & 0.7911      & 0.099         & 1200.8          & 562.6443    & 17.9101      & 2.283      & 0.5023       & 0.1619     & 9.6816           & 5.6134      & 23.9151          & 21.6426         \\
NIN          & 0.6734          & 0.1198        & 1935.4          & 701.686     & 15.6214      & 1.771      & 0.3646       & 0.1333     & 16.9835          & 7.5857      & 74.3866          & 65.9751         \\
ResNet       & 0.7829          & 0.1094        & 1258.9          & 620.7798    & 17.7946      & 2.5017     & 0.4913       & 0.1811     & 8.6024           & 4.7808      & 26.4935          & 24.1246         \\
VGGNet       & \textbf{0.8384} & 0.0955        & \textbf{928.51} & 546.3647    & 19.3142      & 2.8288     & 0.5824       & 0.174      & 7.1588           & 4.3919      & \textbf{14.9142} & 14.0342         \\ \bottomrule
\end{tabular}
\end{table*}

\subsection{Network Architecture Performance}
The six network-bases, described in Sect.~\ref{sec:nets}, were explored to examine if their architectures affect the regression accuracy for this task, and if so, which architecture gives the best performance-to-training-time ratio. All networks were trained using Caffe with the Adam optimizer, max iteration of $200000$, batch size of $64$, learning rate of $0.0001$, momentum of $0.9$, and a learning rate decay of $10\%$ every $20000$ iterations. The exception was the ResNet architecture that utilized a lower batch size of $32$, due to available GPU memory.

Each network is trained with the data set generated by the Euler angle iteration method with Anchor Point labels. Fig.~\ref{fig:train-test-loss} shows the training and testing loss during the training process, where each curve represents the loss of each Anchor Point. 
Table~\ref{tab:net-train-times} shows the number of parameters within each network and the training time for $200000$ iterations.

\begin{table}[htb]
\centering
\caption{Parameter count and training duration for different network architectures}
\label{tab:net-train-times}
\begin{tabular}{@{}ccc@{}}
\toprule
Network          & Number of Parameters & ~~~Training Time~~~ \\ \midrule
CaffeNet         & 32639449             & 1 hr 40 mins     \\
GoogLeNet        & 10253307             & 5 hrs 55 mins    \\
~~Inception v4~~ & 42783001             & 21 hrs 20 mins   \\
NIN              & 6257395              & 16 hrs 30 mins   \\
ResNet           & 66566681             & 25 hrs 40 mins   \\
VGGNet           & 69159385             & 12hrs            \\ \bottomrule
\end{tabular}
\end{table}

We analyze the performance of each network using predefined image similarity metrics, as well as geodesic and Euclidean distances between the predicted and ground truth locations. By comparison to the ground truth, obvious incorrect slice predictions are first discarded (\emph{e.g.}, a slice that is predicted outside the volume). Incorrect slice predictions are defined as possessing a geodesic distance that is more than three scaled Median Absolute Deviations (MAD) from the median.
$
    \text{MAD} = K \cdot \text{median}( \left| x_i - \text{median}(x) \right| ); \quad i = 1, ... , N.
$

\begin{figure}[htb]
  \centering
  \subfloat{\includegraphics[height=1.4cm]{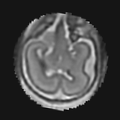}} \hfill
  \subfloat{\includegraphics[height=1.4cm]{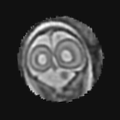}} \hfill
  \subfloat{\includegraphics[height=1.4cm]{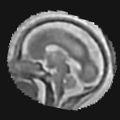}} \hfill
  \subfloat{\includegraphics[height=1.4cm]{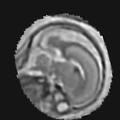}} \hfill
  \subfloat{\includegraphics[height=1.4cm]{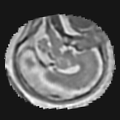}} \hfill
  \subfloat{\includegraphics[height=1.4cm]{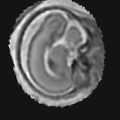}} \hfill \\
  \setcounter{subfigure}{0}
  \subfloat[17]{\includegraphics[height=1.4cm]{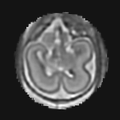}} \hfill
  \subfloat[41]{\includegraphics[height=1.4cm]{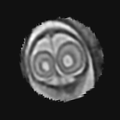}} \hfill
  \subfloat[107]{\includegraphics[height=1.4cm]{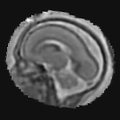}} \hfill
  \subfloat[123]{\includegraphics[height=1.4cm]{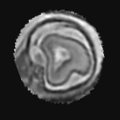}} \hfill
  \subfloat[300]{\includegraphics[height=1.4cm]{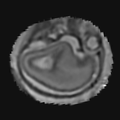}} \hfill
  \subfloat[1637]{\includegraphics[height=1.4cm]{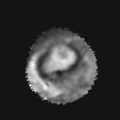}} \hfill
  
  \caption{Top: Validation slices that are presented to the network. Bottom: Slices extracted from the respective fetal volume using parameters predicted by the network. (a) to (f) compares the ground truth slices with predicted slice in order of increasing Geodesic Distance Error.}
  \label{fig:slice-pred}
\end{figure}

Table~\ref{tab:network-performance} shows the average error of 1000 random validation slices that were selected from each of the five test subjects. It can be seen that VGGNet attained the smallest geodesic distance error, as well as best correlation and MSE. This is closely followed by GoogLeNet, which managed to attain the smallest PSNR, SSIM and Average Euclidean Distance error (shows the average prediction error of a single Anchor Point to the corresponding ground truth location). NIN is the worst performing network with much greater errors. For efficiency, GoogLeNet is the ideal choice, as it can attain almost just as good performance as VGGNet in half the training time. If accuracy is of lower importance, CaffeNet may also be used. ResNet, Inception and NIN however take much longer to train and perform worse compared to VGGNet. Fig.~\ref{fig:slice-pred} shows some examples of predictions made by the network.

\subsection{Image Normalization}

The image analysis literature has emphasized the importance of intensity normalization in many domains and it is widely accepted that appropriate normalization is often critical when employing learning based approaches. In this section we report on empirical exploration of the various alternative image normalization strategies considered. Our first experiment involved testing the performance and the effect of Z-score normalization. Slices $\omega_i$ in this data set were generated via the Euler Iterator with Anchor Point labels. Before slice extraction from $\Omega$, each brain volume was Z-score normalized with a mask (\emph{i.e.}, background pixels are not included). The network was trained with the same parameters as defined in Sect.~\ref{sec:impl}, running for a maximum of 280K iterations.

\begin{figure}[htb]
  \centering
  \subfloat[Z-Score normalized data set]{\includegraphics[width=0.24\textwidth]{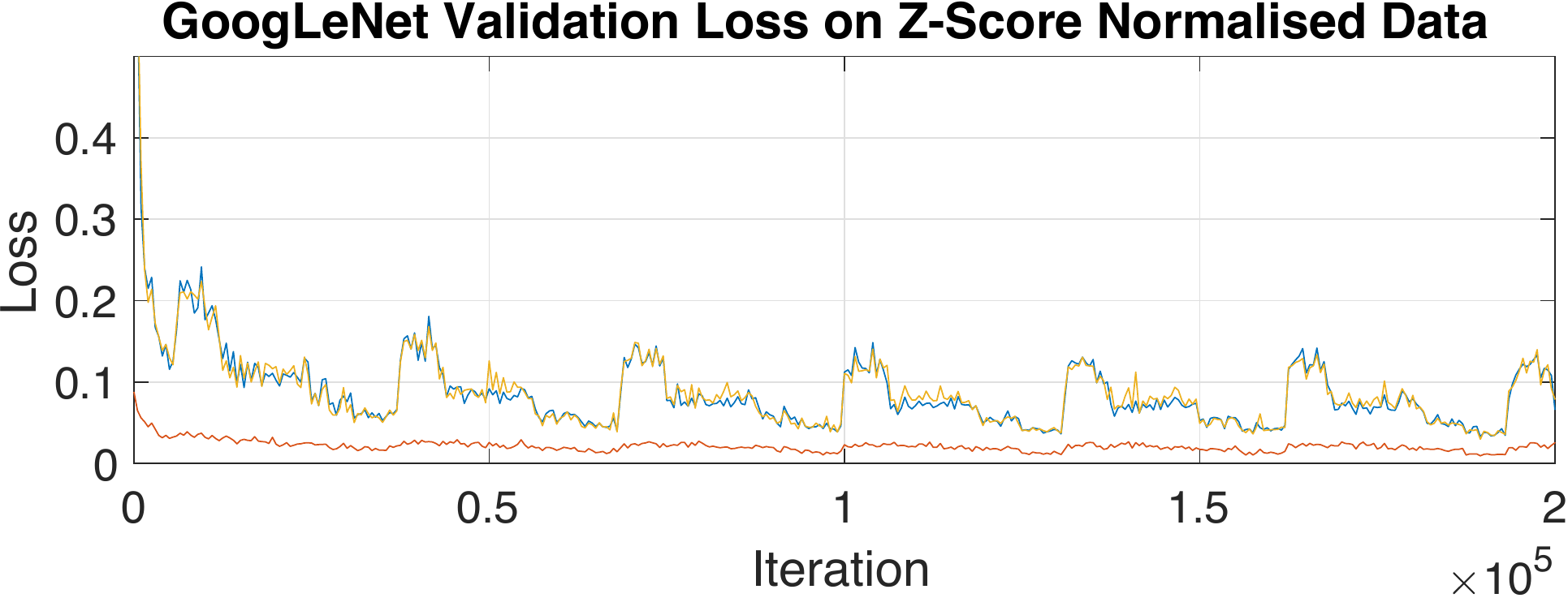}} \hfill
  \subfloat[Histogram matched data set]{\includegraphics[width=0.24\textwidth]{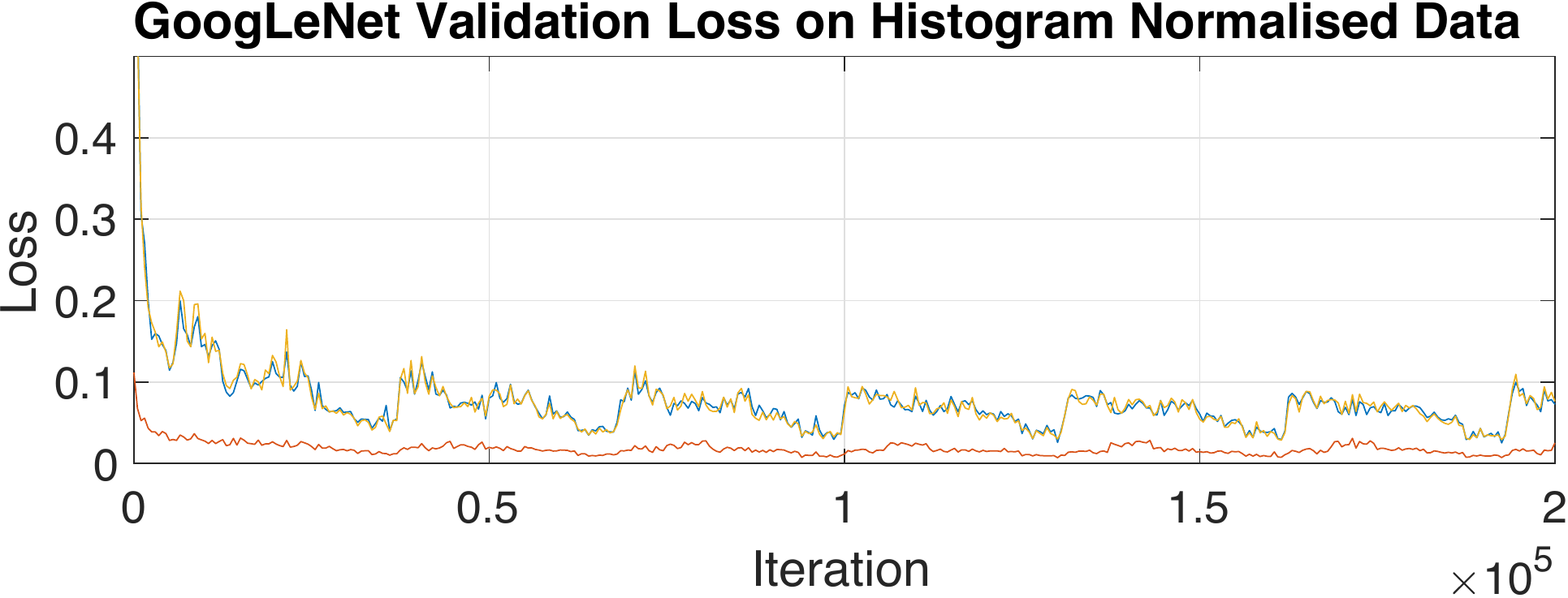}}
  \caption{Validation loss on a network trained via Euler Angle Seed and Anchor Point Loss on (a) Z-Score normalized data and (b) histogram matched data. }
  \label{fig:hist_norm_loss}
\end{figure}

Fig.~\ref{fig:hist_norm_loss}a shows the validation loss during training. The periodicity of the graph indicates over-fitting as it cycles through each subject in the validation database during training. The loss is also slightly higher compared to GoogLeNet trained on the intensity rescaled database as seen in Fig.~\ref{fig:train-test-loss}.

A second normalization experiment alternatively involved taking each $\Omega$ (used for $\omega_i$ generation) matching the intensity profile to a fetal atlas~\cite{brainatlas}. The same procedure was also applied to validation volumes. Slices $\omega_i$ in this data set were also generated via the Euler Iterator with Anchor Point labels. The network is trained for 200K iterations and all other parameters are kept consistent with the previous experimental setup as defined in Sect.~\ref{sec:impl}.


Fig.~\ref{fig:hist_norm_loss}b shows the validation loss for this experiment. It can be seen that the validation loss is lower than that of the Z-score normalization strategy, but higher than that of intensity rescaling. A periodic nature can still be identified, suggesting slight over-fitting. 

A final experiment explores how many different $\Omega$ are needed during training for good generalization. Seven networks were trained with a batch size of 64 and increasing iterations of {40K, 80K, 120K, 160k, 200K, 240K and 280K}. This ensured that every network is trained with 16 epochs, where epoch = iterations $\times$ batch size / data set size. Slices were generated from 4, 8, 12, 16, 20, 24 and 28 fetal volumes, where each volume yields 40K slices via Euler Iterator.

\begin{figure}[htb]
  \centering
  \includegraphics[width=0.48\textwidth]{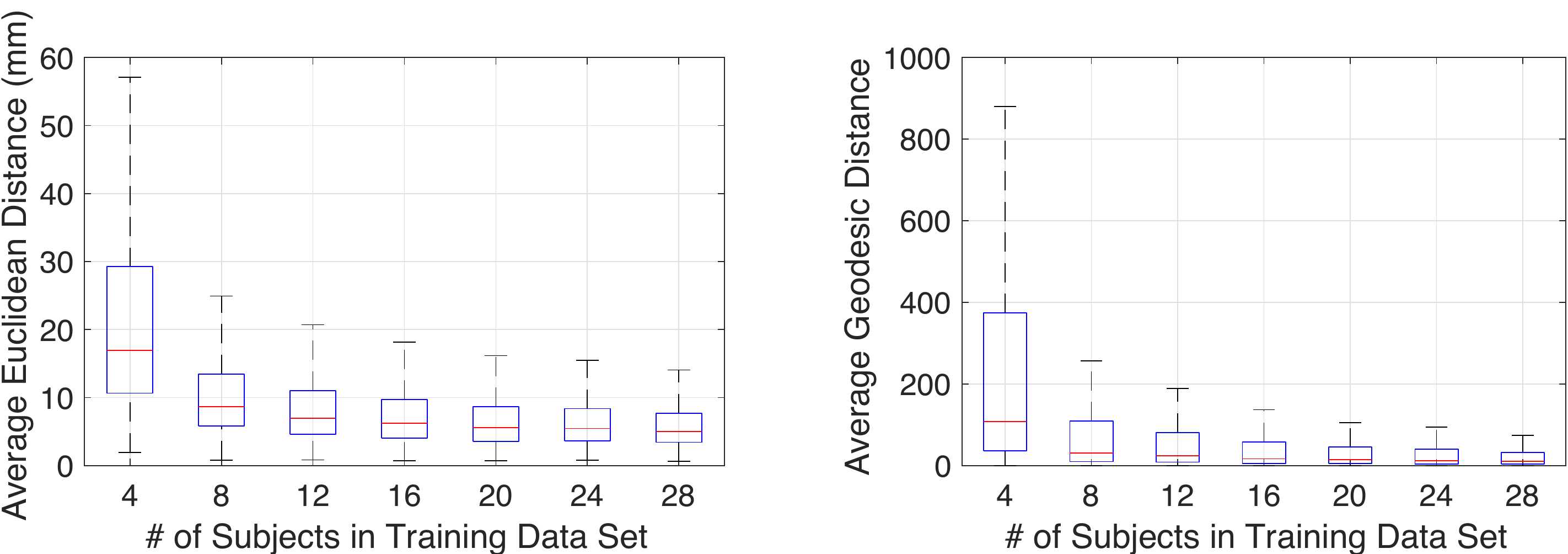}
  \caption{Validation loss on seven networks trained with varying data set sizes. Left: Avg. Euclidean Distance error, Right: Avg. Geodesic Distance error.}
  \label{fig:varying_fetal_subjects}
\end{figure}

Fig.~\ref{fig:varying_fetal_subjects} shows box plots of validation scores of each trained network. The left chart shows Average Euclidean distance error of each Anchor Point, where the right chart shows Geodesic distance error (see Sect.~\ref{sec:metrics}). With 16 to 20 volumes, the Average Euclidean distance error of each Anchor Point (including the quartiles) is already under 10mm. This is within the capture range of recent robust SVR algorithms discussed in Sect.~\ref{sec:relatedworks}. 


\subsection{Regression Labels}

We additionally look to gain an understanding of how our data sets affect performance accuracy as their construction differs, in particular, the different types of label parameterization. Here we make use of the GoogLeNet architecture due to the noted favorable accuracy-speed trade-off that the network possesses. All six data sets adopt the Multi-Loss framework~\cite{Xu:2016:MRD:3024223.3024275}, and uses the same evaluation methodology as before. 5K slices from the validation set (1K per fetal subject) have been randomly selected and passed through the network. Table~\ref{tab:result-accuracy} shows the average errors, for each data set. 

\begin{table}[htb]
\centering
\caption{Average errors for different fetal data sets}
\label{tab:result-accuracy}
\begin{tabular}{@{}cccccc@{}}
\toprule
               & CC             & MSE             & PSNR             & SSIM            & G.E.              \\ 
\midrule
~Anchor Point & \textbf{0.8378} & \textbf{935.0}  & \textbf{19.3564} & \textbf{0.5845} & ~\textbf{15.7504}~ \\
~Eul-Cart     & 0.8148          & 1077.8          & 18.5254          & 0.537           & 18.749             \\
~Quat-Cart    & 0.8199          & 1046.4          & 18.6509          & 0.5448          & 18.1708            \\ 
\bottomrule
\multicolumn{6}{c}{Euler Angle Generation}                                                                  \\
\multicolumn{6}{c}{}                                                                                        \\ 
\toprule
               & CC             & MSE             & PSNR             & SSIM            & G.E.              \\ 
               \midrule
~Anchor Point & \textbf{0.8379} & \textbf{939.6}  & \textbf{19.2297} & \textbf{0.5789} & \textbf{13.6885}   \\
~Eul-Cart     & 0.8078          & 1110            & 18.3034          & 0.5235          & 20.4581            \\
~Quat-Cart    & 0.8147          & 1074.1          & 18.5055          & 0.5383          & 17.7344            \\ 
\bottomrule
\multicolumn{6}{c}{Fibonacci Point Generation}                                                          
\end{tabular}
\end{table}

Here in every metric, Anchor Point labels were able to yield a greater accuracy compared to Euler-Cartesian and Quaternion-Cartesian labels in all test cases. A two-tails independent T-test was conducted to examine the statistical significance as shown in Tab.~\ref{tab:t-test}. As there are 5000 samples in each data set, the DoF is regarded as infinity. The p-values for all tests are therefore infinitesimally small.

\begin{table}[htb]
\centering
\caption{T-test Scores comparing Euler-Cartesian and Quaternion-Cartesian Labels to Anchor Point Labels}
\label{tab:t-test}
\begin{tabular}{@{}cccccc@{}}
\toprule
            & CC          & MSE       & PSNR      & SSIM      & G.E.   ~~ \\ \midrule
~~Eul-Cart  & 10.2973     & 11.0646   & 13.8516   & 12.4705   & 8.0784 ~~ \\
~~Quat-Cart & 8.0766      & 8.8238    & 11.7188   & 10.4969   & 6.6121 ~~ \\ \bottomrule
\multicolumn{6}{c}{Euler Angle Generation}                                \\
\multicolumn{6}{c}{}                                                      \\ \toprule
            & CC          & MSE       & PSNR      & SSIM      & G.E.    ~~\\ \midrule
~~Eul-Cart  & 14.471      & 14.0921   & 16.3974   & 15.248    & 18.9279 ~~\\
~~Quat-Cart & 11.2659     & 11.1481   & 12.5023   & 11.0639   & 12.3679 ~~\\ \bottomrule
\multicolumn{6}{c}{Fibonacci Point Generation}                     
\end{tabular}
\end{table}

\subsection{3D Reconstruction}\label{ref:recon}
We evaluate the proposed pipeline for reconstruction of a 3D MRI fetal brain in order to assess our ability to aid common downstream tasks, that consider accurate input data alignment as a hard prerequisite. 200 synthetically motion corrupted slices are extracted from the validation set in order to initialize SVR~\cite{kainz2015fast}. For all five fetal subjects, we calculate the PSNR between the original and reconstructed 3D volumes using; Gaussian Average and SVR Refinement, see Table~\ref{tab:synthetic-psnr}.

\begin{table}[htb]
\centering
\caption{PSNR of volumes reconstructed from synthetic slices compared to Ground Truth validation volumes}
\label{tab:synthetic-psnr}
\begin{tabular}{@{}cccccc@{}}
\toprule
                     & \#1    & \#2    & \#3    & \#4    & \#5~~    \\ \midrule
~~Gaussian Average~~ & 17.393 & 16.136 & 15.112 & 15.155 & 15.537~~ \\
SVR Refinement       & 18.933 & 19.184 & 18.560 & 18.028 & 20.531~~ \\ \bottomrule
\end{tabular}
\end{table}

\begin{figure*}[htb]
  \centering
  \includegraphics[width=\linewidth]{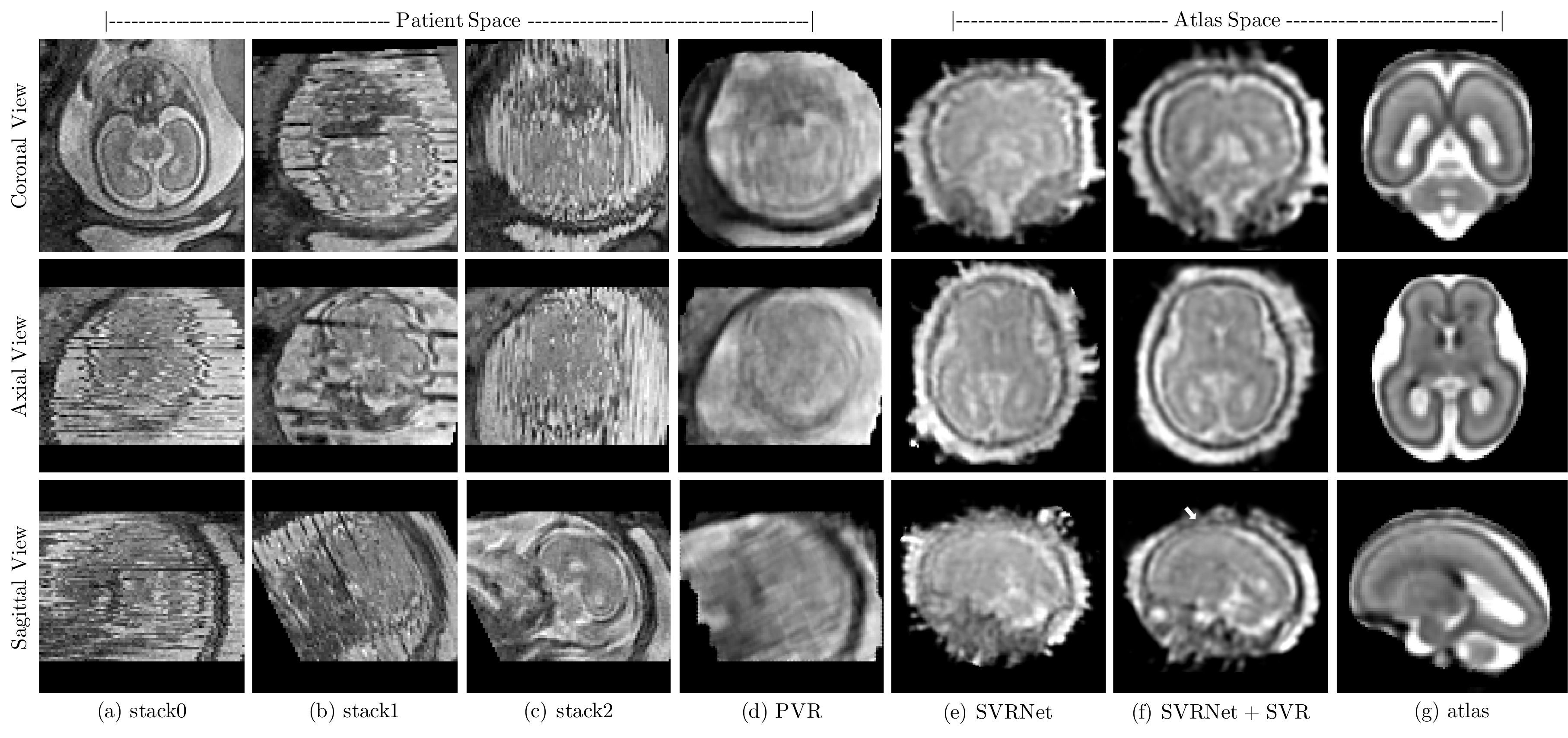}
  \caption{Reconstruction attempt of a fetal brain at approx. 20 weeks GA, from heavily motion-corrupted stacks of T2-weighted ssFSE scans. It was not possible to reconstruct this scan with a significant amount of manual effort by two of the most experienced data scientists in the field using state-of-the-art methods.
  (a), (b) and (c) are 3 example orthogonal input stacks, with scan plane direction in coronal, axial and sagittal respectively. The view direction is initialized to Stack0 as shown in (a). Stacks (b) and (c) are not perfectly orthogonal, as they are taken at different time points and the fetus has moved. (d) Patch to Volume Registration (PVR)~\cite{Alansary2016} with large patches, \emph{i.e.}, improved SVR, using the input stacks directly. (e) Gaussian average of all slices that are predicted and realigned to atlas space by SVRnet. (f) Reconstructed volume after 4 iterations of Slice to Volume Registration (SVR), initialized with slices predicted by SVRnet. The arrow points to an area where insufficient data has been acquired. The fetus moved in scan direction, thus slices are missing in this area necessary for an accurate reconstruction. (g) Training atlas representation of the slices in (e)-(f). Note that (d) is reconstructed in patient space whereas (e) and (f) are reconstructed in atlas space.}
  \label{fig:recon-compare}
\end{figure*}

We further test SVRnet on a case, which our clinical partners dismissed as impossible to reconstruct. Both, extensive manual and automatic reconstruction attempts have failed for this case. With no ground truth to compare to, reconstruction quality can only be validated qualitatively. Fig.~\ref{fig:recon-compare}a, b and c show the raw scan stacks, and the degree of motion corruption. 

\begin{figure}[htb]
  \centering
  \subfloat[\#12]{\includegraphics[height=1.7cm]{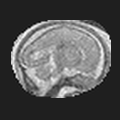}} \hfill
  \subfloat[\#13]{\includegraphics[height=1.7cm]{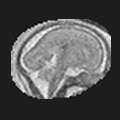}} \hfill
  \subfloat[\#14]{\includegraphics[height=1.7cm]{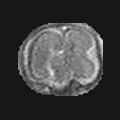}} \hfill
  \subfloat[\#15]{\includegraphics[height=1.7cm]{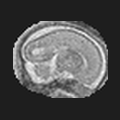}} \hfill
  \subfloat[\#16]{\includegraphics[height=1.7cm]{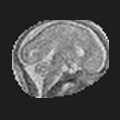}} \hfill 
  \caption{Sequential scan slices from a sagittal image stack of a fetus with extreme motion, it can be observed that the fetus has rotated its head $90^{\circ}$, causing slice \#14 to be a coronal view. 
 }
  \label{fig:extreme-motion}
\end{figure}

In a case like this, excessive motion can cause ambiguity. Fig.~\ref{fig:extreme-motion} shows a sequential sagittal stack of slices where the fetus has turned its head almost $90^{\circ}$, causing a coronal slice to be in a scan stack that is assumed sagittal. This unexpected slice does not fit in the stack, and is normally rejected by robust statistics implemented in SVR algorithms. Rejecting too many slices will cause a lack of scan data, while accepting too many slices will cause a corrupt reconstruction volume as seen in Fig.~\ref{fig:recon-compare}d. Fig.~\ref{fig:recon-compare}d shows a SVR-based reconstruction attempt, using \cite{kainz2015fast}, without SVRnet initialization. 

\begin{figure}[htb]
  \centering
  \subfloat{\includegraphics[height=1.8cm]{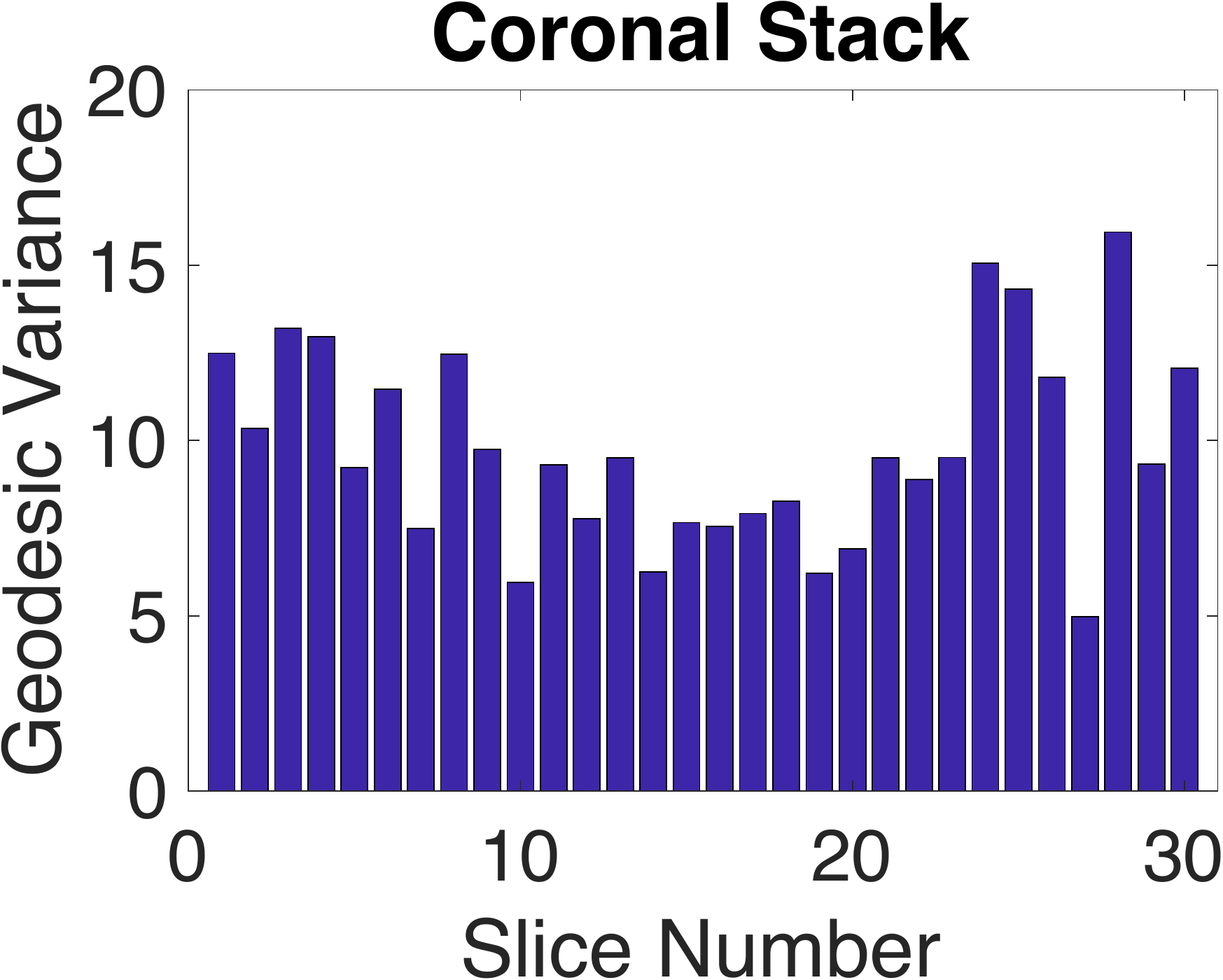}} \hfill
  \subfloat{\includegraphics[height=1.8cm]{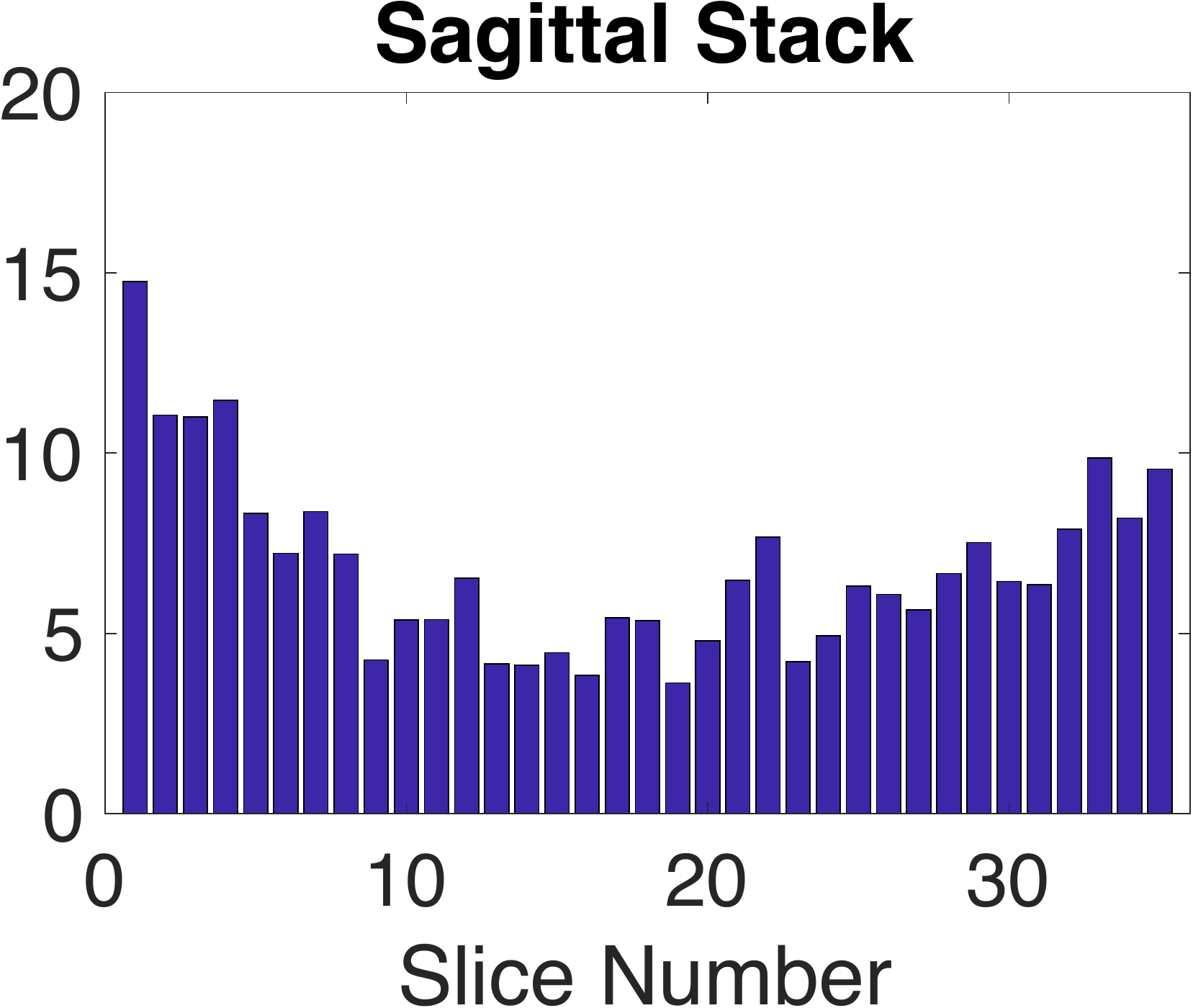}} \hfill
  \subfloat{\includegraphics[height=1.8cm]{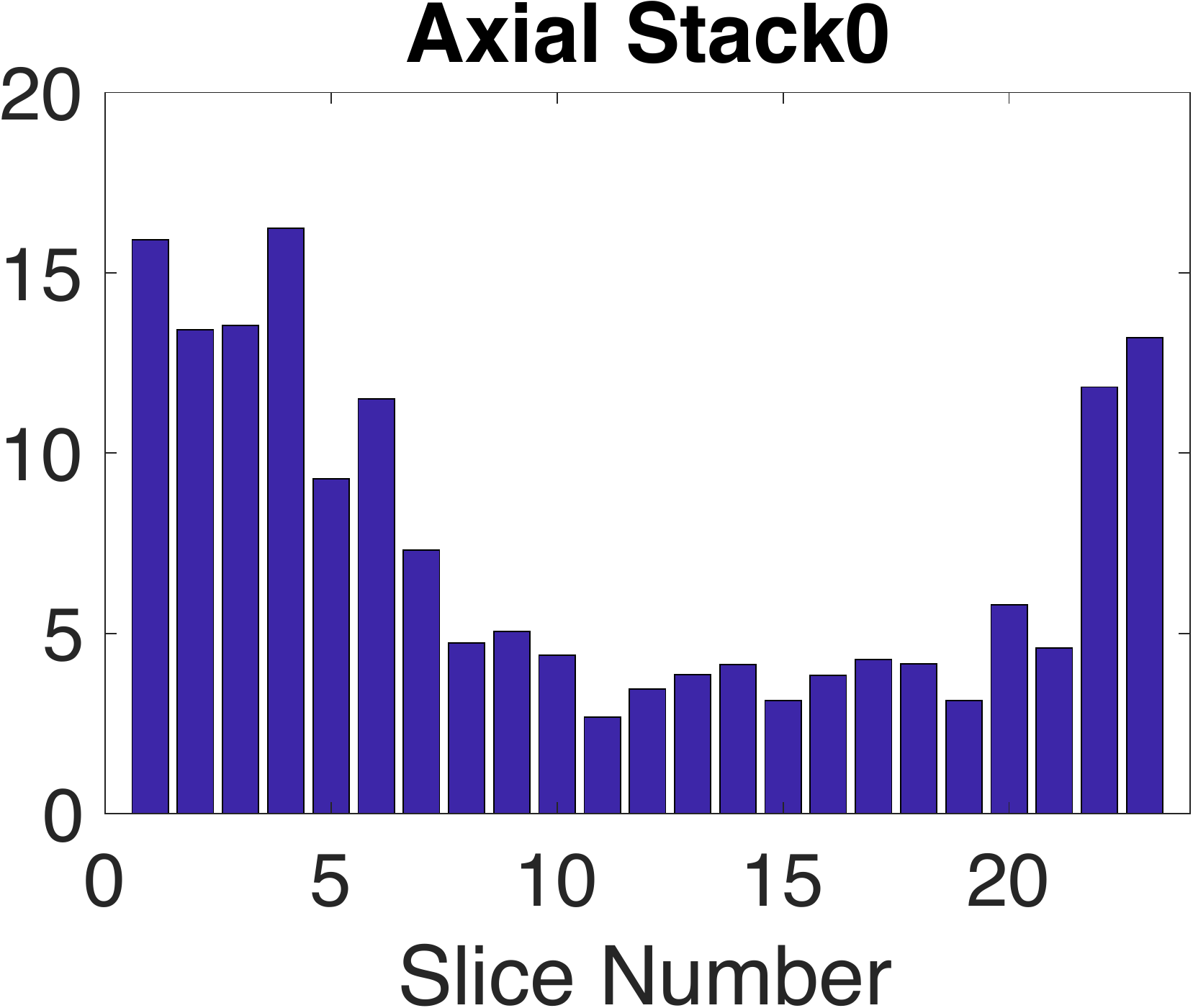}} \hfill
  \subfloat{\includegraphics[height=1.8cm]{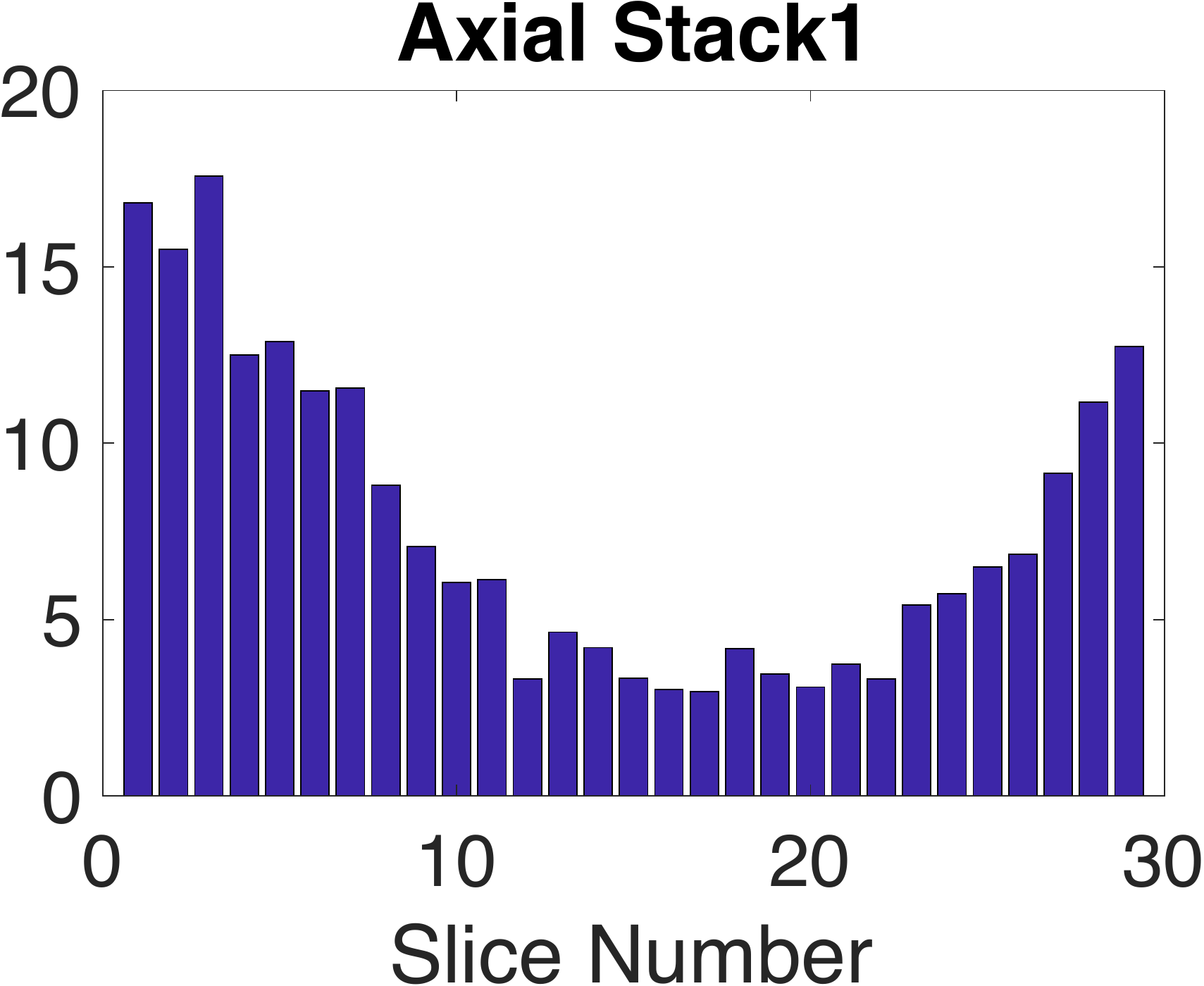}} \hfill 
  \caption{Slice prediction confidence of four heavily motion corrupted orthogonally scanned stacks of slices.}
  \label{fig:pred-conf}
\end{figure}

\begin{figure}[htb]
  \centering
  \subfloat[1.92]{\includegraphics[height=1.1cm]{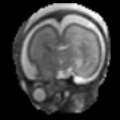}} \hfill
  \subfloat[2.46]{\includegraphics[height=1.1cm]{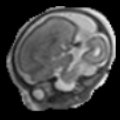}} \hfill
  \subfloat[3.82]{\includegraphics[height=1.1cm]{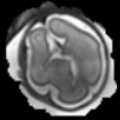}} \hfill
  \subfloat[9.66]{\includegraphics[height=1.1cm]{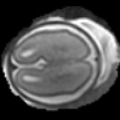}} \hfill 
  \subfloat[29.31]{\includegraphics[height=1.1cm]{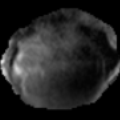}} \hfill
  \subfloat[30.92]{\includegraphics[height=1.1cm]{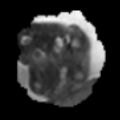}} \hfill
  \subfloat[36.98]{\includegraphics[height=1.1cm]{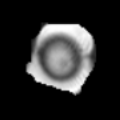}} \hfill
  \subfloat[43.34]{\includegraphics[height=1.1cm]{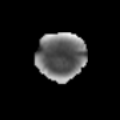}} \hfill
  
  \caption{Monte Carlo predictions of a unitless Geodesic distance variance metric for each slice. A higher number represents greater variance (\emph{i.e.}, the network is less confident). (a)-(d) represent confident predictions. (e)-(h) represent less confident predictions which are discarded for subsequent volume reconstruction.}
  \label{fig:montecarlo_pred}
\end{figure}

Monte Carlo dropout sampling is used for early outlier rejection. Each slice is fed through the network $100$ times. The final prediction is obtained by computing the Riemannian center of mass of all predicted transformations. 
As we have generated slices to train SVRnet from the central portion of the fetal volume, network confidence is lower for boundary slices as shown in Fig.~\ref{fig:pred-conf}. 
Fig.~\ref{fig:montecarlo_pred} shows examples of ``good'' and ``bad'' slices with their corresponding prediction variance.


The decision of whether or not to include a slice in subsequent reconstruction depends on the prediction confidence and the robustness of the chosen reconstruction algorithm for (\textit{3}). Prediction confidence can be thresholded and  
if the reconstruction algorithm is very robust, like~\cite{kainz2015fast}, we can make multiple predictions per slice and let the reconstruction algorithm handle further outlier rejection, which allows for a greater margin of error (see Fig.~\ref{fig:recon-compare}). 

Experimentally we find, for the data sets utilized in this study, a geodesic variance of approx. 10 allows for the reliable distinction between slices useful for subsequent volume reconstruction tasks (confident network predictions) and those which may be discarded (less confident predictions).

Fig.~\ref{fig:montecarlo_pred} (e) and (f) are slices that suffered from signal loss. Fig.~\ref{fig:montecarlo_pred} (g) and (h) are edge case slices where the image plane has only minimal intersection with the brain surface. Such instances make for a high degree of ambiguity in true image plane location. Such cases also prove highly challenging for experienced practitioners without additional information. 

\section{Discussion}
In this paper, we show that a learning based approach (SVRnet) is able to greatly increase the capture range for 2D/3D image registration, and can provide a robust initialization for scans with extreme inter-slice motion corruption. 
The accuracy of the network predictions is influenced by efficient and novel parameterization of labels and loss functions. In particular we explore the effect of utilizing parameterizations that do not lie on a Euclidean manifold. Notably, Euler- and Quaternion-Cartesian labels attain similar levels of performance yet, with a unique parameterization combined with the introduction of Anchor Points, we can further increase the attainable accuracy. 

We evaluated six different architectures with fixed hyperparameter configurations, achieving satisfactory registration accuracy. We provide additional evidence towards typical cost-benefit trade offs of hyperparameter tuning. For the regression of transformation parameters hyperparameter optimization can be extremely time consuming and computationally expensive, whilst providing little improvement in prediction accuracy. Generating synthetic slices for training is very challenging due to the extensive search through a very large space of parameters (6DoF). We constrain the parameter space by leveraging the effects of the in-plane transformations, based on the center-aligned content as a result of organ localization.

SVRnet needs to be retrained for different organs, use case scenarios or modalities (\emph{e.g.}, MRI field strengths, T1, T2, X-Ray exposure, etc.). This can be particularly problematic without organ atlases, or existing 3D reference volumes. 
We have been able to obtain one addition raw 3T scan, and have found that our model, which was trained on 1.5T images, was also able to successfully predict transformation parameters for 3T images. However, further experimentation is necessary to validate the intra-modality robustness of SVRnet. Re-training/transfer learning for each new modality is advised to achieve a maximum of prediction accuracy.

SVRnet requires test images to be formatted in the same way as training, this includes identical intensity ranges, spacing and translation offset removal when pre-processing 3D volumes. 
Our method is not restricted with respect to the used imaging modality and scenario, as seen in \cite{Hou2017} where the network is trained on DRR images as well as whole thorax phantoms. This is valuable for 3D to 2D alignment as the whole volume can be aligned to individual 2D slices.

For the current implementation, SVRnet focuses on rigid transformations. For cases where non-rigid deformation of organs between slices is expected, Patch to Volume (PVR) Reconstruction \cite{Alansary2016} can be used as final volume reconstruction step. For such cases SVRnet will still be able to predict the approximate location of individual 2D slices in canonical 3D space while PVR will handle non-rigid deformations. 

\lastrev{SVRnet uses Euclidean Distance as the primary loss function to regress on Anchor Point labels. It is also possible to use image metrics, such as ones shortlisted in Section~\ref{sec:eval}, as a distance metric if the chosen metric is differentiable. However, metrics like~\cite{Pambrun2015IQALimitations} are intended to be used primarily on natural and not medical images, which means that using such approaches could yield little performance gain.} 


Through experimentation, we find that choices related to slice generation method do not greatly affect prediction performance. In opposition to this, the parameterization strategy representing regression ground truth labels holds significant influence over result quality, as shown in Table~\ref{tab:result-accuracy}. We eventually select the Euler angle parameterization due to ease of use and the requirement of fewer training samples relative to the considered alternatives.

For cases with little motion, SVRnet may be less effective due to the training data sampling interval. For example, we iterate slice rotation in 18$^\circ$ steps for the Euler iterator seed, and 2mm steps in $T_z$. This step size can be intuitively interpreted as the resolution of the network. If the motion corruption that is present is smaller than this interval then the prediction error may introduce a higher slice offset than originally present in the scan. The purpose of SVRnet is not to compete against traditional SVR methods. It is used for the many cases where raw scan volumes are corrupted with motion offsets that are larger than those correctable with traditional SVR methods, where a significant amount of manual intervention would be required. 
As a desirable side-effect, SVRnet predicts slice orientations in canonical atlas co-ordinates, which is not the case for SVR methods.


Another issue, difficult even for human experts, is determining left-right asymmetry of a given slice without additional information. To tackle this issue, oversampling and capturing lots of slices during scan time can allow a greater margin for mis-predicted slices. Robust Statistics~\cite{kainz2015fast} are able to reject slices predicted in the wrong hemisphere. 

The proposed method can also be formulated as a classification task, where each rigid transformation can be quantized as a class. Quantizing the permutations used in our experiments would result in $40K$ to $50K$ classes. With only $28$ fetal examples per class, this will lead to a high class-imbalance introducing new difficulties in training. Reducing the number of classes, however, will decrease the prediction resolution.


\section{Conclusion}
SVRnet is able to predict slice transformations relative to a canonical atlas co-ordinate system, using only the intensity information in the image. This allows motion compensation for highly motion corrupted scans, \emph{e.g.}, MRI scans of very young fetuses. It allows the incorporation of any images that have been acquired during examination, thus relaxing the requirement for temporal scan-plane proximity.

We have evaluated a wide range of state-of-the-art and popular network architectures to examine their performance on prediction accuracy. We found that VGGNet, in our experiments, attained the smallest regression errors. However, GoogLeNet is more efficient to train for repeating experiments. It can achieve similar results to VGGNet with half the training time, whilst occupying $85\%$ less memory space. 

Our work leverages the computational framework to do statistics on SE(3) Lie groups, performing Bayesian Inference and Monte Carlo dropout sampling on the rigid transformation predictions of the network. This approach can also be beneficial in other applications such as~\cite{Kendall_2015_ICCV}, where CNNs are trained to produce output transformations that are not in Euclidean space. We have shown that by calculating geodesic distances of rigid 3D transformations on a non-Euclidean manifold provides means to assess the predicted transformation parameters more accurately. This paves the way to propagate uncertainty downstream in a pipeline that uses the network output to perform other tasks.
  
\lastrev{
\section{Acknowledgements}
We thank the volunteers, the radiographers J. Allsop and M. Fox, MITK~\cite{wolf2005medical}, IRTK (CC public license from IXICO Ltd.), the NIHR Biomedical Research Center at GSTT. 
This work was supported by the Wellcome Trust IEH Award [102431, iFIND]; ERC 319456; NVIDIA (GPU donations); EPSRC EP/N024494/1.
Data access only in line with the informed consent of the participants, subject to approval by the project ethics board and under a formal Data Sharing Agreement.
The views expressed are those of the author(s) and not necessarily those of the NHS, the NIHR or the Department of Health.}
 
%

%

\ifCLASSOPTIONcaptionsoff
  \newpage
\fi



\bibliographystyle{IEEEtran}
\bibliography{IEEEabrv,references}
\end{document}